\PassOptionsToPackage{hyperfootnotes=false}{hyperref}
\documentclass[11pt,a4paper]{article}
\IfFileExists{lmodern.sty}{\usepackage[T1]{fontenc}\usepackage{lmodern}}{}
\usepackage[margin=2cm]{geometry}
\usepackage[authoryear,sort]{natbib}
\usepackage{amsmath,amssymb}
\usepackage{booktabs,tabularx,array}
\usepackage{microtype}
\usepackage{xurl}
\usepackage{placeins}
\usepackage{graphicx}
\usepackage{capt-of}
\usepackage{ragged2e}
\usepackage[labelfont=bf,font=small]{caption}
\usepackage{hyperref}
\graphicspath{{figs/}{./}}
\hypersetup{colorlinks=true,linkcolor=black,citecolor=black,urlcolor=black,
  pdftitle={Aggregate accuracy conceals concentrated temporal vulnerability in a spiking speech classifier},
  pdfauthor={Ismail Can Dikmen}}

\newcommand{\ind}{\mathbf{1}}
\newcommand{\scores}{\mathbf{s}}
\newcommand{\inputtensor}{\mathbf{X}}
\newcommand{\unitcell}{\mathbf{E}}
\title{Aggregate accuracy conceals concentrated temporal vulnerability in a spiking speech classifier}

\author{%
  {\.{I}}smail Can Dikmen\thanks{Corresponding author. E-mail: \href{mailto:can.dikmen@istinye.edu.tr}{\texttt{can.dikmen@istinye.edu.tr}}. ORCID: \href{https://orcid.org/0000-0002-7747-7777}{0000-0002-7747-7777}.}\\[4pt]
  {\normalsize Department of Electrical and Electronics Engineering, \.{I}stinye University,}\\
  {\normalsize Sar\i yer, \.{I}stanbul, 34396, T\"{u}rkiye}}

\date{}

\begin{document}
\maketitle

\begin{abstract}
Aggregate accuracy cannot reveal which utterances are locally vulnerable or how internal activity changes when labels remain stable. We retain every prediction for 725,070 adjacent-bin, one-count changes around 100 validation utterances of a frozen SpikeSCR-based classifier. The canonical native-horizon GPU singleton path reaches 86.0836\% validation accuracy. Equal-source expected accuracy under a uniformly chosen neighbor rises from 84.00\% to 84.54\%, although 13 of 84 initially correct sources admit adverse neighbors. Five sources carry 93.21\% of adverse moves. Margin-guided and surrogate-gradient searches miss sparse cases at fixed query budgets; a post hoc gradient prefix finds all 13 at 73.45\% of the census of initially correct sources. Source-matched traces and clean-state interventions distinguish internal change from harmful direction and recoverability. Two count-readout replicas have similar validation accuracies but a 5.21-fold class-change gap concentrated in four sources. Controlled query/key source isolation eliminates 531 batch-order label changes and restores bitwise order invariance across all 9,981 validation score vectors. Isolated batch predictions reproduce padding-matched singleton labels on 9,980 of 9,981 inputs. Paired CPU/GPU replay of all 25,820 class-changing neighbors gives 99.8993\% label agreement and preserves all 13 vulnerable sources and their fixed witnesses. Complete source-conditioned maps distinguish vulnerability incidence, concentration, internal change, and execution dependence that aggregate accuracy leaves unresolved.
\end{abstract}


\section{Introduction}
\label{sec:introduction}

Consider two football matches played by the same teams and the same players. Both can end with the same score even though the movements, attacks, and course of play differ. The score records the outcome, not the sequence that produced it. A classifier has the same limitation: a preserved label does not establish that its internal processing was unchanged. Across many inputs, a further ambiguity appears. An average score can improve because some mistakes are corrected while other, previously correct decisions become wrong.

This ambiguity matters when evaluating the temporal robustness of spiking neural networks (SNNs). Their time-dependent states and spike communication allow an input timing change to alter later activity \citep{eshraghian2023}. A constructive result also shows that preserving one layer's spike counts while changing its delays can yield non-monotone task accuracy \citep{yan2026falcon}. An evaluation based only on average accuracy can combine adverse and corrective transitions into a favorable net result. A flip rate combines harmful, beneficial, and wrong-to-different-wrong transitions. Even a successful attack leaves unanswered how many other utterances are vulnerable and how densely harmful perturbations populate each local neighborhood. These distinctions determine what an evaluation actually establishes about the model.

The Spiking Speech Commands (SSC) dataset provides a concrete setting because its inputs are event streams generated by a cochlear model \citep{cramer2022}. Surrogate-gradient learning supports the training of these networks despite non-differentiable spike generation \citep{neftci2019}. Adaptive recurrent neurons, learned delays, and global-local processing have advanced spiking speech classification \citep{yin2021,bittar2022,hammouamri2024,deckers2024,wang2026spikescr}. Recent evidence also shows that SSC and Spiking Heidelberg Digits contain useful information beyond spike counts \citep{yu2026}. The question addressed here is how to measure the local consequences of that temporal dependence without allowing averages or incomplete coverage to hide its distribution.

Existing approaches provide essential but different information. Adversarial methods optimize harmful perturbations \citep{liang2023,buchel2022,lun2025}, including count-preserving retiming \citep{yuyi2026}. Formal and direct-search methods determine whether a bounded counterexample exists \citep{banerjee2023,seong2024,mukhoty2024}. Observational explanations characterize internal activity \citep{kim2021}, while activation replacement tests the effect of supplying a selected internal state \citep{meng2022,zhang2024patching}. The missing link in the present setting is a retained local outcome map that connects the distribution of adverse and corrective decisions to internal responses and matched interventions.

We construct that map for a frozen SpikeSCR-based classifier. For each of 100 label-stratified SSC validation utterances, we enumerate every distinct input obtained by transferring one integrated count unit to an adjacent 5 ms bin within the same transformed feature. All 725,070 predictions are retained. The map measures what uniform, margin-guided, and surrogate-gradient searches discover at fixed candidate-query budgets and, crucially, what they leave unseen. Internal traces and clean-activation replacement connect the outcomes to the activity and recoverability of the same runs.

The resulting contrast is direct: equal-source expected accuracy under this finite-neighborhood experiment rises from 84.00\% to 84.54\%, while 13 of the 84 initially correct utterances admit an adverse neighbor. Five sources carry 93.21\% of harmful moves. Low-margin allocation improves expected discovery, but a rare vulnerable source lies outside its selected quartile; a fixed gradient ranking misses different rare cases. A complementary audit of two separately seeded count-readout SNN replicas, using the same utterance identities at their native input resolution, separates spike changes lost under temporal aggregation from decision changes. Nearly equal validation accuracies in the replicas coexist with markedly different class-change counts and a reversed ordering of fully robust correct sources. Controlled execution experiments additionally identify cross-source coupling at the q/k LIF stages as the mechanism of batch-order dependence, while paired CPU/GPU singleton replay preserves the 13-source temporal-vulnerability finding.

The contributions are fourfold:
\begin{enumerate}
\item We retain every outcome of a defined temporal neighborhood and distinguish net correctness, adverse-source incidence, candidate density, and concentration. Counting identities and finite-sampling probabilities are analysis tools for this measured map.
\item We show that favorable mean correctness coexists with highly uneven adverse vulnerability in SpikeSCR, and evaluate uniform, margin-guided, and single- and multi-target gradient discovery at matched candidate-query budgets. The map identifies the sources and rare moves each search misses.
\item We separate decision, representation, and trajectory changes through source-matched SpikeSCR activations, matched branch-state interventions, and exact spike-count readout analysis in two additional SNN replicas. The paired replica comparison shows why class-change load and source-level robustness need separate reporting.
\item We identify cross-source q/k LIF coupling as the mechanism of batch-order dependence in the pinned implementation. Source isolation removes the measured order effect and recovers 9,980 of 9,981 padding-matched singleton labels; paired CPU/GPU replay confirms persistence of the original vulnerable sources and fixed witnesses.
\end{enumerate}

Throughout, a complete audit means complete coverage of the stated discrete neighborhood for the specified inputs, checkpoint, and execution path. This bounded definition makes coverage verifiable and the resulting comparisons reproducible.

\section{Related work}
\label{sec:related}

\subsection{Temporal perturbations and attacks in SNNs}

Early studies examined how input encoding and spiking neuron dynamics
affect adversarial robustness. Sharmin et al. investigated the effects
of discrete input encoding and neuronal leak under gradient-based
attacks \citep{sharmin2020}. Subsequent work developed training procedures
to improve robustness, including crafted input noise in HIRE-SNN
\citep{kundu2021} and regularized adversarial training in SNN-RAT
\citep{ding2022}. These studies concern image-classification settings;
their findings do not by themselves establish robustness to
count-preserving temporal changes in event-based speech.

Later studies considered event-based vision, spike-compatible
gradients, sparse attacks, and limitations of surrogate-gradient
evaluation \citep{buchel2022,liang2023,lun2025,xu2025}.
Wang et al. further show that the surrogate-gradient formulation used
during an attack can materially affect the assessment of SNN
robustness \citep{wang2026robustness}. Attack outcomes therefore depend
on the optimization procedure and its numerical choices. Failure
to find a harmful perturbation does not establish that none exists
\citep{athalye2018}. Our audit addresses this coverage question for a
restricted finite neighborhood by evaluating every admissible candidate.

Yu et al. provide the closest perturbation precedent \citep{yuyi2026}. Their spike-retiming attack preserves spike count and amplitude while optimizing time shifts under $B_\infty$, $B_1$, and $B_0$ budgets, capacity-one constraints, and non-overlap rules. Their projected-in-the-loop method searches for strong discrete attacks on event-vision datasets. Our objective is complementary and our operator is not the same threat model. We move one integrated count unit in the transformed SpikeSCR input by exactly one 5 ms bin, keep it in the same feature, allow an already occupied target, and evaluate every member of this explicitly defined neighborhood. Exactness in this paper applies only to that set; it does not imply a solution to the more general retiming problem of Yu et al.

\subsection{Verification and complete local maps}

Formal SNN verification converts neuron dynamics and perturbation constraints into a satisfiability or optimization problem \citep{banerjee2023,mukhoty2024}. Seong et al. also introduce direct counterexample search over a finite temporal shift set \citep{seong2024}. Their procedure can stop when it finds a class-changing witness and returns a robust/not-robust decision. This is an important enumeration precedent. The contribution examined here is a retained outcome map: every candidate prediction is stored, transitions are typed, and source-level rates and concentrations are calculated even after a witness has been found. A geometric account of robustness also appears in the work of
Calaim et al., who study how low-dimensional representations and
constraints on subthreshold voltages preserve computation under
network perturbations \citep{calaim2022}. Their analysis concerns the
geometry of network dynamics. Here, local decision geometry refers
to the distribution of classifier outcomes over explicitly enumerated
input changes around each source utterance.

\subsection{Timing in spiking speech and internal explanation}

Timing sensitivity in spiking speech is not new. Yu et al. show that surrogate-gradient-trained SNNs can use timing beyond spike counts and study SHD and SSC under randomized timing replacement, together with detailed jitter, deletion, and reversal analyses \citep{yu2026}. Their work establishes that temporal information matters at the dataset and model level. We ask a finer local question: which individual adjacent-bin moves cross a frozen model's decision boundary, which preserve it, and how are those outcomes distributed across sources?

Yan et al. prove this count-preserving delay effect for a constructed network and use harmful and corrective prediction counts in Falcon's delay search \citep{yan2026falcon}. Their perturbation changes a layer's emission schedule. Our audit instead enumerates one-count input moves around a trained speech classifier and retains source incidence, local density, and transition concentration.

Time-resolved observational explanations have been developed for SNNs.
For example, the Spike Activation Map of Kim and Panda uses inter-spike
intervals to form visual explanations \citep{kim2021}. Internal
interventions address a different question: whether changing a
representation changes the downstream output. In language models,
Vig et al. use causal mediation analysis to examine the contribution
of internal components \citep{vig2020}, while Meng et al. restore clean
activations in corrupted runs to investigate factual predictions
\citep{meng2022}. Zhang and Nanda show that activation-patching
conclusions can depend on the corruption procedure and evaluation
metric \citep{zhang2024patching}.

We use matched clean-activation replacement to examine the response
to a specified temporal input change. A clean SpikeSCR boundary state
is inserted into the retimed run from the same source, and the
downstream computation is recomputed with fixed weights. Recovery
establishes that the supplied state is sufficient to restore the
decision under that intervention. It does not establish unique
necessity or identify a single module as the origin of the change.
Clean-prediction restoration and ground-truth restoration are reported
separately because the clean prediction can itself be incorrect.

Table~\ref{tab:positioning} places the study alongside these research
lines. The contribution is a joint analysis protocol for a frozen spiking speech classifier: complete enumeration of a stated one-count adjacent-bin neighborhood, retention of every prediction and transition type, source-conditioned summaries, and matched internal-state replacement, released with every record needed to regenerate the reported tables \citep{dikmen2026data}. The complete map quantifies both the offset between corrective and adverse outcomes and the vulnerable sources missed by uniform, margin-guided, and gradient-ranked searches at finite budgets. Traces and matched replacement connect this coverage analysis to the internal response. Temporal perturbation, finite search, and activation replacement provide the methodological foundations; the contribution is the resulting account of unevenly distributed temporal vulnerability.

\begin{table}[!htbp]
\centering
\caption{Relationship to the closest research lines. "Complete" in the final row means every member of the stated finite neighborhood is retained.}
\label{tab:positioning}
\begin{tabularx}{\textwidth}{>{\raggedright\arraybackslash}p{0.18\textwidth}
>{\raggedright\arraybackslash}p{0.23\textwidth}
>{\raggedright\arraybackslash}p{0.24\textwidth}
>{\raggedright\arraybackslash}X}
\toprule
\textbf{Research line} & \textbf{Main question} & \textbf{Typical output} & \textbf{Difference from this study} \\
\midrule
Optimized SNN attacks \citep{liang2023,buchel2022,lun2025,yuyi2026}
& Can an optimizer find a harmful perturbation?
& Attack success, robust accuracy, or a selected adversarial example
& Does not retain every outcome of our collision-permitting one-count SSC neighborhood. \\
\addlinespace[5pt]
Formal or exact search \citep{banerjee2023,seong2024,mukhoty2024}
& Does any bounded counterexample exist?
& SAT/UNSAT, certificate, witness, and runtime
& Direct search may stop at a witness; our output is a full transition-aware local decision map, not a global certificate. \\
\addlinespace[5pt]
Speech timing studies \citep{yu2026}
& Does timing carry useful information, and how does aggregate accuracy change under temporal disruption?
& Dataset-level accuracy curves under replacement, jitter, deletion, or reversal
& Does not enumerate and retain every adjacent-bin neighbor for each audited source. \\
\addlinespace[5pt]
SNN explanation \citep{kim2021}
& Where and when is internal spiking activity salient?
& Observational heatmaps
& Does not perform matched internal-state replacement under retiming. \\
\addlinespace[5pt]
Internal interventions outside SNNs
\citep{vig2020,meng2022,zhang2024patching}
& How do interventions on internal representations affect model outputs?
& Mediation effects, clean-state restoration, and sensitivity to patching choices
& Provides methodological precedents for internal intervention;
the present audit applies matched replacement to retimed spiking speech inputs. \\
\addlinespace[5pt]
\textbf{This study}
& What is the complete local outcome map, how does activity change, and which clean boundaries are sufficient for recovery?
& Every prediction and transition type, source vulnerability, informed-search coverage, internal trace, and matched replacement outcome
& Retains the complete declared neighborhood and source-conditioned outcome distributions. \\
\bottomrule
\end{tabularx}
\end{table}

\section{Materials and methods}
\label{sec:methods}

\subsection{Study design and separation of data roles}

The main SpikeSCR study followed model development and checkpoint selection with three locked evaluation stages. First, the selected SpikeSCR-based classifier was evaluated once on the official SSC test split. This stage measured external classification performance only. Second, the local temporal neighborhood of 100 fixed validation utterances was enumerated under exact singleton inference. Third, internal activations were recorded for the same candidates, and clean activations were inserted at selected model boundaries. The second and third stages were validation-only analyses. Their code prohibited access to the official test data and read only the already committed benchmark scalars.

The checkpoint, preprocessing, definitions of the test metrics, validation panel, and perturbation rule were fixed before the test file was opened. No test prediction was used to select a checkpoint, change a setting, exclude an example, or define a perturbation. The official test split was not perturbed. A complementary count-readout audit uses two separately seeded SNN replicas and the same 100 validation utterance identities (Section~\ref{sec:count_readout_methods}). Those replicas have their own earlier training and test-evaluation history; their local audit does not access test examples. The full-validation singleton runs, padding decomposition, q/k source-isolation controls, paired CPU/GPU candidate replay, and clean-input gradient ranking were conducted after the local audit to diagnose its execution and compare search policies. Sampling expectations and paired replica summaries were calculated from retained validation records. None of these analyses reopened the official test split.

\begin{figure}[!htbp]
\centering
\includegraphics[width=\textwidth]{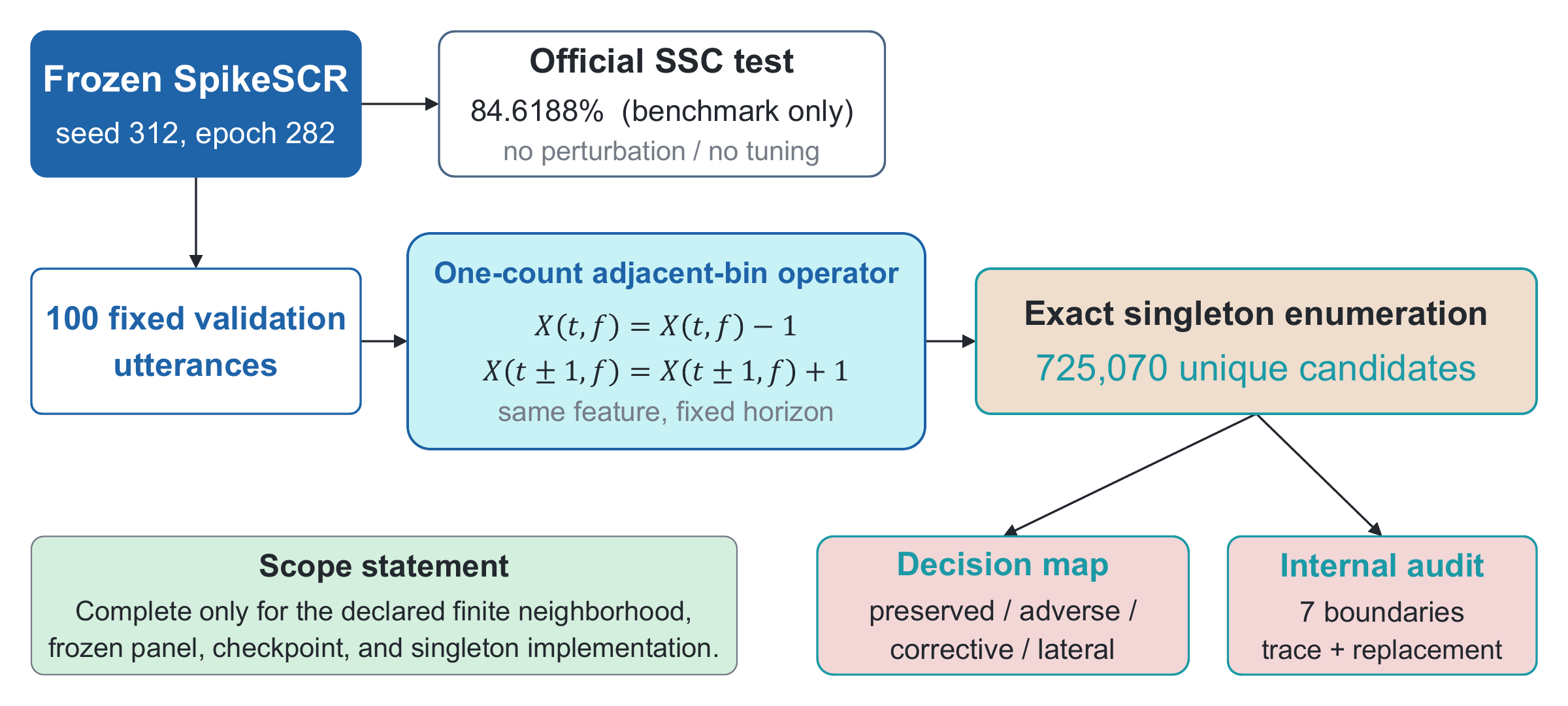}
\caption{Study workflow and separation of data roles. The frozen SpikeSCR checkpoint was evaluated once on the official SSC test split for benchmark measurement. The temporal audit used 100 fixed validation utterances. One integrated count unit was transferred to an adjacent 5 ms bin within the same transformed feature, and every valid candidate was evaluated with native-horizon CUDA singleton inference. For each candidate, one count unit was moved to either the preceding or the following bin, provided that the target remained within the source horizon. The resulting candidates were used for the decision map and the internal trace and replacement analyses. Completeness applies only to the declared finite neighborhood, frozen panel, checkpoint, and CUDA singleton implementation.}
\label{fig:audit_workflow}
\end{figure}

The separation between the benchmark and the validation audit is summarized in Fig. \ref{fig:audit_workflow}. The official test branch ends at the benchmark result. All perturbation, geometry, tracing, and replacement analyses start from the frozen validation panel.

\subsection{Dataset, input representation, and classifier}

SSC contains 35 spoken-command classes recorded as asynchronous events \citep{cramer2022}. We used the official training, validation, and test partitions. The validation and test partitions contain 9,981 and 20,382 utterances, respectively.

The public SpikeSCR implementation was pinned to commit \texttt{095f418f53b3} \citep{wang2026spikescr}; the complete commit identifier and source digest are retained in the result manifest. The network retains the public SpikeSCR architecture and has 3,302,400 trainable parameters, or 3,302,416 parameters including the fixed rotary-embedding parameters. It contains a spiking embedding stem followed by two global and local blocks. The inherited SpikeSCR blocks combine global attention with local convolution. The attention path extends the spiking self-attention of Spikformer \citep{zhou2023spikformer} with rotary position encoding.
The local path adapts the Conformer convolution module \citep{gulati2020} to spiking computation. The global-local arrangement draws in part on Spike-driven Transformer V2 \citep{yao2024spikev2}. The hidden dimension is 256, the attention module has 16 heads, and the local kernel size is 31.

Raw SSC unit identifiers range from 0 to 699. Five adjacent units were combined into one transformed feature, giving 140 input features. Yan et al. use the same five-channel grouping for SSC, with 10 ms bins rather than the 5 ms bins used here \citep{yan2026falcon}. Event times were measured relative to the first event and accumulated in 5 ms bins. The transformed input for utterance $i$ was a non-negative integer matrix
\begin{equation}
    \inputtensor_i \in \mathbb{N}_0^{H_i \times 140},
\end{equation}
where $H_i$ is the utterance-specific transformed horizon. The nominal one-second horizon is 200 steps. The 100 audited sources had horizons between 111 and 200 steps.

The inherited SpikeSCR readout sums time-step softmax values. For time-step logits $\mathbf{z}_t$, its class score and prediction were
\begin{equation}
    s_c(\inputtensor)=\sum_t \operatorname{softmax}(\mathbf{z}_t)_c,
    \qquad
    \hat y(\inputtensor)=\arg\max_c s_c(\inputtensor).
    \label{eq:score}
\end{equation}

The archived official-validation accuracy of 86.3340\% and official-test accuracy of 84.6188\% belong to the frozen public-release batch-256 GPU evaluation path. Every canonical local decision, trace, and activation-replacement result uses the same weights with the unmodified network at semantic and computational batch size one. The canonical decision census was executed on CUDA at each source's native horizon. A later complete CPU singleton validation run classified 8,591/9,981 inputs correctly (86.0735\%). Controlled GPU replays compared native-horizon singletons, singletons padded to their batch-group horizon, ascending-order batch-256 inputs, and those same padded batches in reversed source order. Appendix~\ref{app:execution} documents their paired contrasts and the additional mechanism controls. The batch-256 test value remains specific to its recorded execution. Earlier checks found layout dependence in the public q/k path; a rejected samplewise executor failed rare exact-score checks, requiring exhaustive reconstruction with unmodified singleton inference.

\subsection{Model development and checkpoint selection}
\label{sec:model_development}

The model retains the public SpikeSCR architecture. It was trained from randomly initialized weights for 300 epochs with seed 312, 5 ms bins, batch size 256, AdamW learning rate $5\times10^{-3}$ and weight decay $10^{-2}$, cosine scheduling with $T_{\max}=40$, and the public time-neuron masking augmentation. The epoch-282 checkpoint achieved the highest official-validation accuracy in that run, 86.3340\%, and is used throughout the main audit. Model, optimizer, scheduler, and random-number-generator states were retained.

Compatibility adaptations replaced the CuPy neuron backend with the PyTorch backend, wrapped multi-step batch normalization without changing its parameters or running statistics, set the missing \texttt{use\_ln} field to \texttt{false}, and called AdamW directly. Dependency versions, source hashes, and preprocessing checks were recorded. Earlier DCLS screens, readout comparisons, and learning-rate continuations are documented in Supplementary Section~S1. They do not substitute checkpoints in this audit.

\subsection{Frozen validation panel}

The local audit used 100 distinct official-validation utterances. They were sampled without replacement by ground-truth label, with two or three sources from each of the 35 classes.  The selection seed was 906605967 and source selection used only labels and source identifiers. Clean predictions, margins and perturbation outcomes were not available to the selection rule.

This design gives similar representation to each class. Consequently, the mean of the 100 source-level rates describes the fixed label-quota panel. It is not a prevalence-weighted estimate for all 9,981 validation utterances. Under exact singleton inference, 84 of the 100 panel sources were classified correctly before perturbation.

\subsection{Finite adjacent-bin neighborhood}

For every occupied cell $(t,f)$, one integrated count unit was moved to the preceding or following 5 ms bin when that target remained inside the source horizon. For direction $d\in\{-1,+1\}$, a candidate was
\begin{equation}
    \inputtensor_i^{(t,f,d)}
    =
    \inputtensor_i
    -\unitcell_{t,f}
    +\unitcell_{t+d,f},
    \label{eq:operator}
\end{equation}
subject to $X_{i,t,f}>0$ and $0\leq t+d<H_i$. The transformed feature $f$, source horizon $H_i$, and total integrated count were unchanged. The target cell could already be occupied. This is the smallest nonzero displacement on the model's 5 ms count grid at fixed feature and horizon. Its value is a controlled, fully enumerable local stress test of the representation: it distinguishes inputs that differ in one integrated temporal count while leaving the total count unchanged. It is not a physical jitter model, a perceptual-equivalence claim, or a move of one identified raw 700-channel event. Classical spike-train distances provide useful context for this operator. The Victor-Purpura metric assigns costs to spike insertion,
deletion, and temporal displacement \citep{victor1997}; a displacement
of duration $|\Delta t|$ has an elementary edit cost of
$q_{\mathrm{VP}}|\Delta t|$, where $q_{\mathrm{VP}}$ is the chosen
temporal cost parameter. The van Rossum distance instead compares
filtered spike trains at a selected time scale \citep{vanrossum2001}.
Our neighborhood is defined directly on the transformed count tensor
by Eq.~\eqref{eq:operator}. It is not defined as a ball under either
metric, and no equivalence to a unit-distance raw-event perturbation
is assumed.

One unique tensor candidate was retained for each valid occupied-cell/direction pair. If $X_{i,t,f}=m$, that candidate represents $m$ interchangeable choices among events integrated into the same cell. We therefore report two denominators. The \emph{unique-candidate} denominator counts the resulting tensor once. The \emph{event-weighted} denominator assigns multiplicity $m=X_{i,t,f}$. The unique-candidate analysis is primary because it measures distinct neighboring model inputs. Event-weighted results are reported separately and are never mixed with unique-candidate counts.

Every valid candidate in Eq.~\eqref{eq:operator} was evaluated. If $O_i$ is the number of occupied input cells and $B_i$ is the number of occupied cells in the first and last bins, the unique-neighbor count is $N_i=2O_i-B_i$. Each occupied cell supplies two directions except at the horizon boundaries; distinct valid moves produce distinct signed difference tensors. This gives a direct coverage check independent of the network predictions.

\subsection{Decision outcomes and source-level quantities}

Let $y_i$ be the ground-truth class, $\hat y_i$ the clean prediction, and $\hat y_{ij}$ the prediction for candidate $j$ of source $i$. Each candidate was assigned to exactly one outcome:
\begin{equation}
\begin{aligned}
\text{class-preserved:}\quad& \hat y_{ij}=\hat y_i,\\
\text{adverse:}\quad& \hat y_i=y_i,\quad \hat y_{ij}\neq y_i,\\
\text{corrective:}\quad& \hat y_i\neq y_i,\quad \hat y_{ij}=y_i,\\
\text{lateral:}\quad& \hat y_i\neq y_i,\quad \hat y_{ij}\neq y_i,\\
& \hat y_{ij}\neq\hat y_i.
\end{aligned}
\label{eq:transitions}
\end{equation}

A source was called class-change-sensitive if at least one candidate changed its clean prediction. An initially correct source was called adverse-sensitive if at least one candidate made it incorrect. Candidate-pooled rates and equal-source mean rates answer different questions and are reported separately.

The clean margin was
\begin{equation}
    M_i=s_{\hat y_i}(\inputtensor_i)-
    \max_{c\neq\hat y_i}s_c(\inputtensor_i).
    \label{eq:margin}
\end{equation}
Because $s_c$ sums softmax mass over the source horizon, $M_i$ is measured in summed probability mass and can vary with $H_i$. The association between clean margin and adverse-candidate rate was evaluated only among the 84 initially correct sources. Dividing $M_i$ by $H_i$ supplied a sensitivity analysis for margin-guided search.

\subsection{Accuracy decomposition and finite-sampling coverage}
\label{sec:coverage_methods}

Let $w_{ij}\geq0$ be fixed weights over source-neighbor pairs, with $\sum_{i,j}w_{ij}=1$. The matched clean and neighboring accuracies are
\begin{equation}
\begin{aligned}
\mathrm{Acc}_0^w&=\sum_{i,j}w_{ij}\ind[\hat y_i=y_i],\\
\mathrm{Acc}_{\mathcal N}^w&=\sum_{i,j}w_{ij}\ind[\hat y_{ij}=y_i].
\end{aligned}
\end{equation}
Write $P_w(A)$ and $P_w(C)$ for the weighted adverse and corrective masses. The transition partition gives the exact identity
\begin{equation}
\mathrm{Acc}_{\mathcal N}^w-\mathrm{Acc}_0^w=P_w(C)-P_w(A).
\label{eq:accuracy_decomposition}
\end{equation}
An adverse transition contributes $-1$ to correctness, a corrective transition contributes $+1$, and preserved or lateral outcomes contribute zero. Thus a positive net change does not exclude adverse neighbors. We use $w_{ij}=1/(100N_i)$ for equal-source reporting and $w_{ij}=1/\sum_iN_i$ for unique-candidate reporting. The clean baseline must use the same weights as its neighboring counterpart.

The complete map also supplies a reference for limited uniform sampling. Let $A_i$ be the adverse count among the $N_i$ distinct neighbors of an initially correct source. Sampling $m_i$ distinct neighbors uniformly without replacement detects at least one adverse outcome with probability
\begin{equation}
p_i(m_i)=1-\frac{\binom{N_i-A_i}{m_i}}{\binom{N_i}{m_i}}.
\label{eq:detection_probability}
\end{equation}
The numerator counts samples containing no adverse member. It is zero when $m_i>N_i-A_i$. For a common budget $m$, we set $m_i=\min(m,N_i)$ and sum $p_i(m_i)$ across initially correct sources to obtain the expected number of vulnerable sources discovered. These are exact hypergeometric calculations for the stated sampling design. Equations~\eqref{eq:accuracy_decomposition} and \eqref{eq:detection_probability} are standard counting identities used to interpret the measured map.

We also evaluated fixed informed policies over the same 84 initially correct sources and total candidate-query budgets. The lowest-quartile policy allocated queries equally among the 21 smallest clean margins. A mixture assigned half of the queries uniformly across all 84 sources and half across those 21, using capped proportional allocation with deterministic largest-remainder rounding. Candidate sampling remained uniform without replacement within each allocated source. Source margins, neighborhood sizes, and IDs set the allocation; candidate outcomes entered only the subsequent exact coverage calculation. The policies and budgets were specified before the numerical sweep, but after the original audit had revealed its outcomes. Results are retrospective comparisons on this panel, not held-out policy performance. To show the effect of the selected cutoff, we repeated hard-focus allocation for the lowest 10\%, 25\%, and 50\% of sources, rounding the selected source count upward to 9, 21, and 42. Both raw and horizon-normalized margins were used. Once all focused neighborhoods were exhausted, remaining queries were distributed uniformly among the other sources. This sweep measures cutoff sensitivity; it does not independently validate the policy choice.

A second informed baseline ranked legal moves separately within each source using clean-input surrogate gradients. For clean predicted class $\hat y_i$ and clean runner-up $r_i$, the single-target objective was $s_{r_i}(X_i)-s_{\hat y_i}(X_i)$; a move from $(t,f)$ to $(t+d,f)$ received the first-order score $g_{t+d,f}-g_{t,f}$, where $g$ is the input gradient of that objective. One backward pass was run for each of the 100 panel sources, although only the 84 initially correct sources enter the discovery comparison.

We additionally computed $g^{(c)}=\nabla_X[s_c(X_i)-s_{\hat y_i}(X_i)]$ for all 34 rival classes, giving 3,400 backward passes over the full panel. Two rankings were fixed: the maximum directional gain $\max_{c\ne\hat y_i}[g^{(c)}_{t+d,f}-g^{(c)}_{t,f}]$, and the maximum predicted target margin, which adds each rival's clean score difference $s_c(X_i)-s_{\hat y_i}(X_i)$ before taking that maximum. The second accounts for the rival's initial distance from the clean decision. Both use the same 34 gradients and require 2,856 backwards for the 84 eligible sources. Ties use the stored candidate index. Candidate outcomes enter only after rankings have been saved.

Batch normalization and dropout remained in evaluation mode while surrogate LIF gradients were enabled. For all 100 sources, the surrogate-mode hard-spike forwards and scores agreed bit-for-bit with normal evaluation, and the runner-up gradient within the multi-target calculation exactly reproduced the earlier single-target gradient. Fixed top-$k$ rankings were compared with the retained canonical outcomes. Candidate-query counts exclude the extra clean forward and backward work; these are first-order rankings rather than iterative optimized attacks. Fresh singleton forwards independently checked the first adverse witnesses of both multi-target policies.

\subsection{Internal activation traces}

Internal output tensors were captured at seven declared boundaries: the spiking embedding stem; the attention, local conformer-convolution, and complete-block outputs of block 1; and the corresponding three outputs of block 2. Attention and local outputs were the branch-level boundaries of scientific interest. The stem and complete-block outputs were also used as positive controls for activation replacement: supplying the complete clean
output at these boundaries is expected to reproduce the clean downstream computation.

For boundary $b$, let $\mathbf{a}_{ib}$ and $\mathbf{a}_{ijb}$ denote the flattened clean and candidate activation tensors. The block outputs include residual sums and are not binary spike tensors; a relative $L_2$ value for these outputs must not be interpreted as a spike-flip fraction. The primary trace measure was relative $L_2$ change,
\begin{equation}
    D_{ijb}=
    \frac{\lVert\mathbf{a}_{ijb}-\mathbf{a}_{ib}\rVert_2}
         {\max(\lVert\mathbf{a}_{ib}\rVert_2,10^{-12})}.
    \label{eq:relative_l2}
\end{equation}
Mean absolute change, root-mean-square change, cosine distance, activity-state change, and signed mean change were retained as secondary descriptors. To describe the magnitude of the final-block response, we also summarize the stored relative $L_2$ measurements by transition type using medians, quartiles, and equal-source means. These post hoc descriptive summaries were calculated from the existing trace records without additional model inference. Exact-zero changes were retained, and quartiles used linear interpolation. For each outcome, the equal-source mean averages the within-source means over sources containing at least one candidate of that outcome. We also summarize the distribution of within-source medians for the preserved outcome. Quartiles describe the observed distribution and are not confidence intervals; no additional hypothesis tests were introduced.

Traces were obtained for all 725,070 candidates. Only traces produced by the unmodified network with computational batch size one were eligible. Previously computed singleton traces for 51,560 candidates were reused after their output scores were shown to equal the canonical scores; the remaining 673,510 traces were recomputed.

We additionally relate final-block activation divergence to the stored score geometry. Define the ground-truth margin at an input $X$ as $\gamma_i(X)=s_{y_i}(X)-\max_{c\ne y_i}s_c(X)$ and let $\Delta\gamma_{ij}=\gamma_i(X_{ij})-\gamma_i(X_i)$. For each source, we calculate Spearman correlations between final-block $D_{ijb}$ and three descriptors: score-vector displacement $\|\scores(X_{ij})-\scores(X_i)\|_2$, absolute margin change $|\Delta\gamma_{ij}|$, and signed margin change $\Delta\gamma_{ij}$. We report the median of the 100 source-specific correlations. A sensitivity analysis excludes exact-zero final-block responses, because unchanged activations and scores otherwise share zero ranks. These descriptive correlations are retrospective analyses of the joined canonical records; no new confirmatory hypothesis tests are introduced.

\subsection{Matched clean-activation replacement}

Each of the 25,820 class-changing candidates was subjected to a matched activation-replacement intervention. At one declared boundary, the candidate output tensor was replaced by the clean output tensor from the same source. Only the selected boundary tensor was manually replaced. Model weights and downstream operators were held fixed, and the downstream computation was recomputed from the patched state. Activations downstream of the intervention could therefore change. Seven interventions were evaluated for every class-changing candidate, giving 180,740 patched forward passes.

For an adverse candidate, restoration of the clean prediction is also restoration of the ground-truth class. For corrective and lateral candidates, restoration of the clean prediction is not a repair measure because the clean prediction is incorrect. Clean-prediction and ground-truth restoration were therefore retained separately.

Replacement at the stem and complete block outputs served as positive controls. The attention and local boundaries were the main branch-level interventions. A successful replacement shows that the matched clean state at that boundary is sufficient to restore the downstream decision under this intervention. It does not show that the boundary, branch, or any individual neuron is uniquely necessary, nor does it identify the first point at which the perturbation became causal.

\subsection{Complementary two-seed count-readout audit}
\label{sec:count_readout_methods}

To examine temporal aggregation directly in a second architecture, we include the completed prospective audit of two separately seeded, frozen SSC replicas, A and B (seeds 341969035 and 554720563, respectively). Their training provenance specifies a fixed architecture and 33 training epochs per seed, with no validation- or test-based seed exclusion. Each network has two causal convolutional branches of 192 spiking units, kernel size 5 and dilations 1 and 4, followed by 320 recurrent LIF units with subtractive reset and a linear count readout. Inputs use 700 cochlear channels and 100 bins of 10 ms. This representation is native to these replicas; its quantitative rates are reported separately from the 5 ms, 140-feature SpikeSCR audit.

The same 100 utterance identities were selected using labels and source identifiers before the audit outcomes were calculated. Each replica was evaluated on all 1,297,705 unique, same-channel, adjacent-bin one-count moves at its fixed input horizon. The audit records the complete 704-unit binary spike trace, final-layer counts, logits, and predictions. Identical-input replays, count conservation, directional partitions, and tie-aware prediction reconstruction were checked. The completed records contain 10,592 null replays and no recorded integrity violations.

Let $S\in\{0,1\}^{100\times320}$ be the final-layer spike tensor and let $P$ collect its global count and four consecutive 25-bin window counts. The readout is
\begin{equation}
C=\operatorname{vec}(PS),\qquad z=W(C/d)+b,\qquad d=1\;\mathrm{s},
\label{eq:count_readout}
\end{equation}
where $C$ denotes the flattened 1,600-component integer count vector. The logits are computed in float32; integer exactness refers to $C$. A final-layer change $\Delta S\ne0$ satisfying $P\Delta S=0$ leaves the readout features and, under the recorded fixed execution, the logits unchanged. This gives a direct test of temporal changes removed by aggregation.

Four exclusive outcomes describe the trace-readout-decision sequence: T0, the complete recorded spike trace is unchanged; T1, the trace changes but $C$ is unchanged; T2, $C$ changes but the predicted class is unchanged; and T3, the class changes. Within T1, we distinguish changes that reach the final spike layer from early-layer changes that disappear before it. The registered batch-384 validation replays classified 7,214/9,981 (72.2773\%) and 7,239/9,981 (72.5278\%) examples correctly. These replays differed by $+1$ and $-2$ correctly classified examples from the earlier training records; these differences are retained in the provenance. The local audit uses its recorded frozen execution and does not claim bitwise reproduction of those earlier dataset-level runs. Both replicas had prior official-test evaluations during their original training replication; no test examples were accessed by the local audit or the present reanalysis.

\subsection{Statistical analysis and integrity controls}

Candidate-level totals describe the enumerated neighborhood. Candidates from the same utterance share the input signal, label, clean state, and model parameters, and were therefore not treated as independent replicates. The source utterance was the unit of resampling and hypothesis testing. This choice addresses the dependence problem discussed for nested observations by \citet{saravanan2020}. Because each source neighborhood was exhaustively evaluated, bootstrap samples resampled sources rather than individual candidates within a source.

The class-change rate was first calculated within each source and then averaged across sources. Bootstrap intervals followed the resampling framework described by Efron and Tibshirani~\citeyearpar{efron1993}. For initially correct sources, the mean adverse rate and its 95\% interval were obtained with 10,000 label-stratified source-bootstrap samples. The margin association used Spearman's $\rho$ and a one-sided negative-association test with 10,000 within-label permutations.

Three source-level geometry contrasts were tested: preceding minus following-bin class-change rate, late minus early time-tertile rate, and high minus low feature-quartile rate. For internal traces, a transition-specific source contrast subtracted that source's mean class-preserved relative $L_2$ change from its mean relative $L_2$ change for the selected transition. Two-sided source sign-flip tests and source-bootstrap intervals used 10,000 repetitions. Benjamini-Hochberg false-discovery-rate correction \citep{benjamini1995} was applied separately to the geometry, transition-versus-preserved trace, attention-versus-local trace, and attention-versus-local patch families. Adjusted values are denoted by $q$.

For the two count-readout replicas, the retrospective paired comparison aligns the same 100 source identities and the same native neighbor sets. A 20,000-replicate bootstrap resampled sources within ground-truth labels and kept each source's two model outcomes paired. The resulting intervals quantify source variation conditional on this panel and these frozen models; they do not estimate variation over training seeds. The common-clean-correct cohort is reported separately because adverse transitions are defined only for sources initially classified correctly by the model in question.

Monte Carlo permutation and sign-flip $p$ values were calculated as $p=(b+1)/(B+1)$, with $B=10{,}000$. For the one-sided negative-association test, $b$ counts permuted correlations less than or equal to the observed correlation. For the two-sided sign-flip tests, it counts absolute resampled mean contrasts greater than or equal to the absolute observed mean contrast.
The smallest attainable value is therefore $1/10{,}001$, reported as $p=0.00010$ after rounding. This value marks the finite resampling resolution. Sign-flip tests used fixed, test-specific random seeds. Benjamini-Hochberg correction was applied to the unrounded $p$ values; distinct unadjusted values can consequently yield identical adjusted $q$ values.

All canonical neighborhood and internal analyses used deterministic float32 singleton inference with automatic mixed precision and TF32 disabled. Spiking state was reset between forward passes. Input-count conservation, source horizon, candidate uniqueness, score/prediction agreement, transition partitions, and archive manifests were checked before aggregation. A final singleton reconstruction corrected ten candidate predictions across six sources relative to the rejected adapted approximation. The complete CPU validation run matched all 100 canonical clean panel predictions, with maximum score discrepancy $5.72\times10^{-6}$. Native GPU singleton replay reproduced all 100 canonical clean score vectors bit-for-bit and passed ten identical-input replay checks. Identities, labels, horizons, and input hashes agreed across the complete validation references. The later paired device run reproduced all 25,820 canonical class-changing predictions on GPU and checked every corresponding CPU outcome. Contracts bind the executed code, checkpoint, dependencies, input hashes, and runtime to the retained results. Only the unmodified singleton outcomes define the canonical local map.

Numerical results were produced by executing the recorded scripts on frozen inputs, model states, and retained outcomes. Reproduction materials \citep{dikmen2026data} include script hashes, dependency versions, checkpoint digests, per-source gradients, fixed rankings, and executable coverage checks. The all-target extension was checked against the earlier runner-up gradients and same-runtime hard-spike replays; its first adverse witnesses were also checked by new singleton forwards.

\subsection{Execution mechanism and paired device controls}
\label{sec:execution_methods}

All execution controls used the frozen checkpoint, validation cache, pinned model source, evaluation mode, deterministic float32, disabled automatic mixed precision and TF32, and state resets before and after each forward pass. The public q/k path directly reshapes a rotary-encoded $[B,h,T,d]$ tensor to $[T,B,N]$ before its second LIF stage, where $N=hd$. Here \texttt{q\_lif2} and \texttt{k\_lif2} are the second LIF stages in the query and key branches, following rotary encoding. The direct reshape can place values from different sources along one membrane lane's declared time axis. A diagnostic intervention isolated sources only at \texttt{q\_lif2} and \texttt{k\_lif2}, retaining each source's original singleton head/time convention. Batch-size-one calls bypassed the intervention; unchanged panel scores therefore establish preservation of that reference condition, rather than validate the nontrivial batch transformation. Validation compared isolated batch execution with all 9,981 padding-matched singleton references and used paired order and duplicate-input controls. Three retrospectively selected pairs from order-sensitive groups were run as $[x,y]$, $[y,x]$, and $[x,x]$, with an identical-input repeat, under original, q-only, k-only, and joint isolation. The population contrast used all 9,981 validation inputs in fixed batch groups under original and reversed order, with and without joint isolation. Appendix~\ref{app:qk_mechanism} gives the layout transformation.

Padding diagnosis reran all 344 sources whose historical singleton label changed under padding and 64 deterministic positive-padding controls. For each of these 408 sources, native scores were compared with the valid-prefix and added-bin components of the padded readout. The added component's total softmax mass was checked against the number of added bins. These selected diagnostic sources do not constitute a new random panel.

The device control reconstructed every one of the 25,820 originally class-changing candidates across 25 sources and ran it as an unmodified native-horizon singleton on CPU and GPU in the same runtime. Both score vectors were retained; transition types were recomputed against that device's clean prediction. The first archived adverse candidate of each of the 13 vulnerable sources was fixed before device outcomes were observed. No source-isolation intervention was active during this replay.

\section{Results}
\label{sec:results}

\subsection{Classifier performance and singleton-batch agreement}

The canonical native-horizon GPU singleton path classifies 8,592/9,981 validation inputs correctly (86.0836\%; Table~\ref{tab:benchmark}). This is the unmodified batch-size-one function used for the local audit. The same checkpoint produced 8,617/9,981 correct predictions (86.3340\%) in archived batch-256 validation. After the experimental contract had been frozen, the batch-256 path was evaluated once on the official SSC test split: 17,247/20,382 correct (84.6188\%). The later CPU singleton control classifies 8,591/9,981 correctly (86.0735\%) and differs from canonical GPU singleton execution on two labels (Table~\ref{tab:execution_controls}). The SpikeSCR study reports 84.47\% for a two-block, 200-step direct-training SSC setting \citep[Table~7]{wang2026spikescr}. Its checkpoint and execution contract are not established by the present controls; the comparison locates our classifier in the model literature rather than measures controlled superiority.

\begin{table}[!htbp]
\centering
\caption{Performance of the frozen SpikeSCR-based checkpoint under separate execution contracts. The test split was evaluated once on the archived batch-256 path; no singleton test evaluation was made.}
\label{tab:benchmark}
\footnotesize
\setlength{\tabcolsep}{2pt}
\begin{tabularx}{\columnwidth}{@{}Xrrrr@{}}
\toprule
\textbf{Split} & \textbf{Correct / $n$} & \textbf{Accuracy} & \textbf{Balanced} & \textbf{Macro-F1} \\
\midrule
Validation, batch 256 & 8,617 / 9,981 & 86.3340\% & 85.2682\% & 85.7325\% \\
Validation,\newline GPU singleton\newline (audit path) & 8,592 / 9,981 & \textbf{86.0836\%} & 84.8246\% & 85.4290\% \\
Validation, CPU singleton & 8,591 / 9,981 & 86.0735\% & 84.8030\% & 85.4092\% \\
Official test, batch 256 & 17,247 / 20,382 & 84.6188\% & 83.7526\% & 84.0844\% \\
\bottomrule
\end{tabularx}
\end{table}

The archived batch-256 and CPU singleton paths differ by 0.2605 percentage points in net accuracy, yet change 617/9,981 predicted labels (6.18\%): 224 batch-correct cases become singleton-incorrect, 198 go in the opposite direction, and 195 are wrong with different predicted classes in both paths. On the fixed 100-source audit panel, 84 clean predictions are correct under singleton execution and 85 under archived batch-256 execution; eight labels differ. Two singleton adverse-sensitive sources are already clean-incorrect under batch-256 execution. The 13/84 local incidence below therefore characterizes the canonical GPU singleton function; the clean panel labels also agree with its CPU replay. The lowest archived batch-256 test recall was 57.8397\% for class 31; its validation-to-test accuracy difference was 1.7152 percentage points.

The controlled GPU study reproduces every archived batch-256 validation label. Native GPU singleton inference attains 8,592/9,981 correct predictions (86.0836\%) and differs from CPU singleton inference on two labels. The score vectors differ numerically for all 9,981 inputs, but the median source-wise maximum absolute difference is $1.91\times10^{-6}$ and the 95th percentile is $3.81\times10^{-6}$; 134 sources exceed $10^{-4}$, with a maximum of 3.0513. Native singleton and batch-256 execution on the same GPU nevertheless differ on 619 labels (6.20\%). Table~\ref{tab:execution_controls} separates these contrasts; disagreement counts are not additive components of the total discrepancy.

Padding native singleton inputs to their batch-group horizons changes 344 labels (3.45\%). Equation~\eqref{eq:score} sums the padded output bins as well as the valid prefix: padding a 111-bin source to 200 bins adds 89 softmax steps to the readout. No labels change among the 4,057 sources receiving no padding, compared with 24/104 (23.08\%) receiving at least 81 added bins (Table~\ref{tab:padding_gap}). The selected 408-source decomposition reproduces every historical native and padded label. Excluding added output bins restores 26 of the 344 changed labels; the other 318 remain changed in the valid-prefix readout. Padding therefore changes valid-prefix processing as well as adding output mass. The extra-output explanation alone is insufficient.

Reversing source order within otherwise unchanged padded batch-256 groups changes 531 labels (5.32\%) and lowers accuracy from 86.3340\% to 85.8631\%. All 9,981 score vectors change, with median source-wise maximum absolute difference 1.3283. Joint q/k source isolation eliminates this order effect: every original and reversed-order score vector is bitwise identical, and both accuracies are 8,589/9,981 (86.0535\%). After isolation, batched inference reproduces padding-matched singleton labels on 9,980 of 9,981 inputs; the remaining difference is one wrong-to-correct transition (Table~\ref{tab:execution_controls}). The three paired controls also lose both order sensitivity and duplicate-input position sensitivity under joint isolation; q-only or k-only isolation leaves score differences. Together, these interventions identify cross-source coupling in the q/k LIF layout as the mechanism of the measured order dependence (Appendix~\ref{app:qk_mechanism}).

\subsection{Complete local decision geometry}

The 100 sources generated 725,070 unique adjacent-bin candidates, or 1,659,158 candidates after event multiplicity was applied. Source neighborhoods ranged from 3,633 to 12,014 unique candidates, with a mean of 7,250.7.

A total of 25,820 unique candidates changed the clean class, corresponding to a pooled rate of 3.5610\%. The equal-source mean class-change rate was 3.9192\%. Table \ref{tab:geometry_outcomes} gives the complete outcome partition.

\begin{table}[!htbp]
\centering
\caption{Complete outcome partition for the specified adjacent-bin neighborhood. Percentages use the corresponding unique or event-weighted denominator.}
\label{tab:geometry_outcomes}
\small
\setlength{\tabcolsep}{3pt}
\renewcommand{\arraystretch}{1.15}
\begin{tabularx}{\columnwidth}{@{}>{\raggedright\arraybackslash}Xrrrr@{}}
\toprule
& \multicolumn{2}{c}{\textbf{Unique}} & \multicolumn{2}{c}{\textbf{Event-weighted}} \\
\cmidrule(lr){2-3}\cmidrule(l){4-5}
\textbf{Outcome} & $n$ & \% & $n$ & \% \\
\midrule
Class-preserved & 699,250 & 96.4390 & 1,601,188 & 96.5061 \\
Any class change & 25,820 & 3.5610 & 57,970 & 3.4939 \\
\hspace{0.4em}Adverse & 5,157 & 0.7112 & 10,424 & 0.6283 \\
\hspace{0.4em}Corrective & 9,031 & 1.2455 & 20,106 & 1.2118 \\
\hspace{0.4em}Lateral & 11,632 & 1.6043 & 27,440 & 1.6539 \\
\bottomrule
\end{tabularx}
\end{table}

Among the class-changing unique candidates, 19.97\% were adverse, 34.98\% corrective, and 45.05\% lateral. A single flip/no-flip rate would treat these harmful, beneficial, and wrong-to-different-wrong transitions as equivalent.

\begin{figure}[!htbp]
\centering
\includegraphics[width=\textwidth]{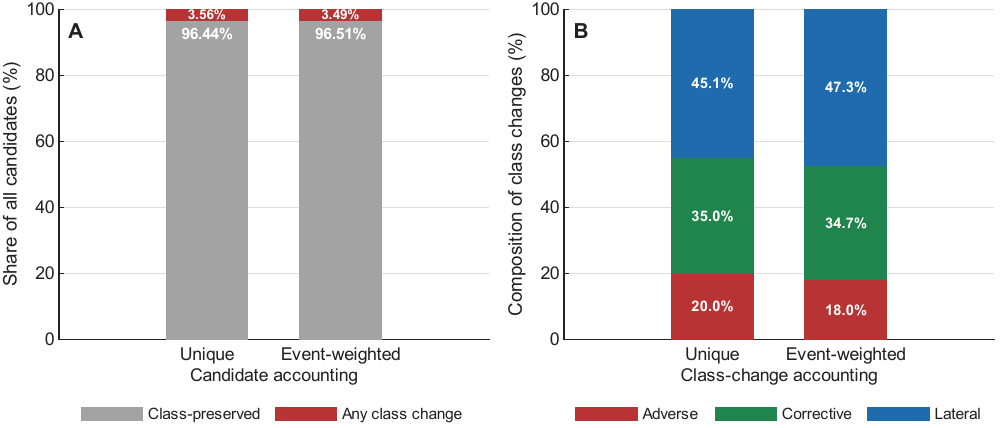}
\caption{Complete outcome partition of the declared adjacent-bin neighborhood. Panel A shows the share of class-preserved and class-changing candidates for unique-candidate and event-weighted accounting. Panel B decomposes only the class-changing candidates into adverse, corrective, and lateral outcomes. Exact counts and percentages are given in Table \ref{tab:geometry_outcomes}.}
\label{fig:outcome_partition}
\end{figure}

Figure \ref{fig:outcome_partition} gives the partition visually. Panel A shows that most candidates preserved the clean class under both accounting rules. Panel B separates the class changes by their relation to the ground truth. This separation is important because a class change is not always an error.

Twenty-five of the 100 sources had at least one class-changing candidate. Among the 84 initially correct sources, 13 were adverse-sensitive, giving a source incidence of 15.48\%. Nine sources contained corrective candidates and eight contained lateral candidates. These source counts can overlap because different candidates from one source may produce different transition types.

The mean adverse rate across initially correct sources was 0.9596\%, with a label-stratified source-bootstrap 95\% interval from 0.2532\% to 1.6978\%. The burden was highly concentrated. Five sources accounted for 93.21\% of all adverse unique candidates, six accounted for 95.77\%, and eight accounted for 99.20\%. The two most affected sources alone accounted for 68.47\%.

Clean margin was negatively associated with adverse rate among the 84 initially correct sources ($\rho=-0.5688$, one-sided within-label permutation $p=0.00010$). Lower margins were associated with higher adverse-candidate rates, although margin did not determine every local outcome.

\begin{figure}[!htbp]
\centering
\includegraphics[width=\textwidth]{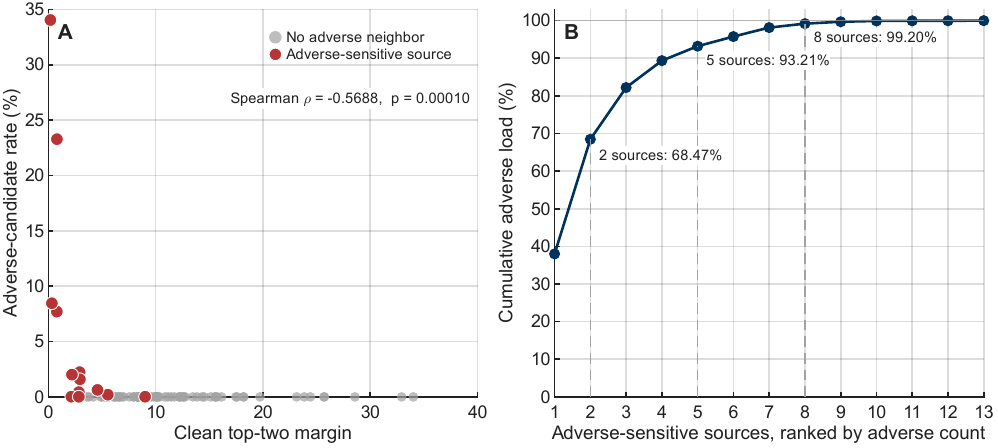}
\caption{Source-level local fragility. Panel A relates the clean top-two margin (summed softmax probability mass across each source's horizon, Eq.~\eqref{eq:margin}) to the adverse-candidate rate for the 84 initially correct sources. Red markers identify the 13 sources with at least one adverse neighbor. The low-margin-quartile membership is unchanged when the margin is divided by horizon. Panel B ranks adverse-sensitive sources by adverse count and gives cumulative load. Two sources account for 68.47\% of adverse candidates, five for 93.21\%, and eight for 99.20\%.}
\label{fig:source_fragility}
\end{figure}

The source-level structure is shown in Fig. \ref{fig:source_fragility}. The margin relationship is visible in panel A, but the scatter also shows that margin alone does not determine local failure. Panel B shows how strongly the adverse load is concentrated among a small number of sources.

No broad temporal or feature-axis contrast survived false-discovery-rate correction. The preceding-minus-following contrast had mean 0.00057 and $q=0.0616$; the high-minus-low feature-quartile contrast had mean $-0.00554$ and $q=0.0616$; and the late-minus-early time-tertile contrast had mean $-0.00051$ and $q=0.7979$. The observed sensitivity was more source-specific than a simple early/late, preceding/following, or low/high-feature rule.

\subsection{Favorable average accuracy hides incomplete vulnerability coverage}
\label{sec:coverage_results}

The transition partition reveals a positive net accuracy change despite the adverse-sensitive sources. Under equal-source weighting, the clean accuracy is 84.00\% and the expected neighboring accuracy is 84.5418\%. Under unique-candidate weighting, the corresponding values are 86.6359\% and 87.1702\%, a gain of 0.5343 percentage points. The higher clean value in the latter calculation reflects the greater weight given to sources with larger neighborhoods. It is not a separate dataset benchmark. Equation~\eqref{eq:accuracy_decomposition} attributes the pooled gain exactly to 9,031 corrective versus 5,157 adverse transitions. Favorable net correctness therefore coexists with adverse neighbors for 15.48\% of initially correct sources.

Uniform sampling misses much of that source incidence: 100 distinct neighbors per initially correct utterance discover an expected 7.61 of 13 adverse-sensitive sources. Informed allocation helps, although its performance varies by source (Table~\ref{tab:sampling_coverage}). At 8,400 total candidate queries, hard focus on the 21 lowest-margin sources discovers an expected 9.42, and a 50/50 uniform-plus-quartile mixture discovers 8.91. The single-target gradient ranking finds seven. Expanding to all 34 rivals finds seven using maximum directional gain and eight using maximum predicted target margin. At 84,000 queries these gradient counts are 11, 10, and 11. The randomized-policy entries are exact finite-population expectations; the gradient entries are deterministic outcomes looked up in the complete map after ranking. The one- and 34-backward rankings incur additional computation beyond their displayed candidate queries.

\begin{table}[!htbp]
\centering
\caption{Discovery among the 13 vulnerable sources within the same 84 singleton-correct sources at fixed total candidate-query budgets. E denotes exact expected discovery for uniform-within-source sampling; D denotes actual discovery by deterministic gradient rankings. Focus thresholds select the 9, 21, and 42 lowest raw-margin sources. The 50/50 allocation mixes uniform and quartile focus. Runner-up uses one backward per source; both all-target rankings use the same 34 backwards per source. Candidate-query budgets exclude this additional work.}
\label{tab:sampling_coverage}
\small
\setlength{\tabcolsep}{5pt}
\begin{tabular}{@{}rrrrrrrrr@{}}
\toprule
\shortstack{Candidate\\queries} & \shortstack{Uniform\\E} & \shortstack{Lowest 10\%\\E} & \shortstack{Lowest 25\%\\E} & \shortstack{Lowest 50\%\\E} & \shortstack{50/50\\E} & \shortstack{Runner-up\\D} & \shortstack{All-target\\gain D} & \shortstack{All-target\\margin D}\\
\midrule
840 & 3.71 & 5.21 & 6.03 & 4.84 & 5.22 & 4 & 4 & 4\\
8,400 & 7.61 & 6.25 & 9.42 & 8.62 & 8.91 & 7 & 7 & 8\\
84,000 & 10.30 & 11.28 & 11.05 & 10.54 & 10.75 & 11 & 10 & 11\\
\bottomrule
\end{tabular}
\par\smallskip\footnotesize At 84,000 queries, the lowest-10\% neighborhoods are exhausted and excess queries are allocated uniformly outside them. Supplementary Section~S2 gives raw and horizon-normalized cutoff results.
\end{table}

The cutoff sweep makes the allocation tradeoff visible. At 8,400 queries, focusing on the lowest 10\%, 25\%, or 50\% gives expected discoveries of 6.25, 9.42, or 8.62; the quartile is not an independently validated optimum. Twelve of the 13 vulnerable sources fall in that quartile, but source 1627 ranks 44th of 84 and has only one adverse move among 6,941, receiving zero hard-quartile queries at these budgets. Horizon normalization keeps quartile membership and expectations unchanged but moves this source to rank 42. Normalized half-focus therefore includes all 13 vulnerable sources, yet at 84,000 queries still discovers only an expected 10.83 and has a 2.15\% probability of finding all 13.

The sparse tail also survives the gradient extensions. The single-target ranking reaches source 1627's adverse move at rank 781 but reaches those of sources 9282 and 1534 only at ranks 4,509 and 5,612. A common post hoc prefix of 5,612 finds all 13 using 461,380 candidate queries, or 73.45\% of the 628,171-candidate census of initially correct sources. The capped count is lower than $5,612\times84=471,408$ because 14 neighborhoods are shorter. The analogous full-discovery prefixes cost 598,437 queries (95.27\%) for all-target gain and 461,870 (73.53\%) for all-target predicted margin. Enumeration is therefore not required merely to find every witness in this map; it additionally determines every source's adverse density and all transition types. For context, a 95\% chance of finding the individually rarest uniform-sampling cases requires thousands of draws, whereas the most affected source needs only eight.

The runner-up and all-target-gain rankings do not exceed uniform expected discovery at 8,400 queries. Adding clean target margins improves that deterministic count from seven to eight, but does not close the sparse-tail gap at the larger budget. This is evidence about these frozen first-order rankings, consistent with the evaluation sensitivity to surrogate gradients reported by \citet{lun2025} and \citet{wang2026robustness}; it does not establish failure of iterative attacks or of other surrogate formulations.

\subsection{Stable decisions often concealed internal change}

A preserved class did not usually imply an unchanged internal computation. Of the 699,250 class-preserved candidates, 617,941 (88.3720\%) had a nonzero relative $L_2$ change at the output of block 2. Across all class-preserved candidates, including those with zero change, the candidate-pooled mean block-2 relative $L_2$ change was 0.5752. All adverse, corrective, and lateral candidates had nonzero block-2 change.

The distribution of preserved responses confirms that the reported mean is not driven solely by a few large changes. Including exact-zero responses, the pooled median was 0.6464, with first and third quartiles 0.6130 and 0.6726. The equal-source mean was 0.5719. Across the 100 within-source medians, the median was 0.6549, with quartiles 0.6322 and 0.6728. Table \ref{tab:activation_distributions} gives descriptive summaries for all four outcomes. Their eligible source sets differ, so comparisons between its rows should be distinguished from the matched contrasts below.

\begin{table}[!htbp]
\centering
\caption{Descriptive final-block relative $L_2$ activation changes calculated from the stored canonical trace records. All candidates, including exact-zero changes, are retained. Brackets contain first and third quartiles, not confidence intervals. The equal-source mean includes only sources containing the specified outcome.}
\label{tab:activation_distributions}
\begin{tabular}{@{}lrrcr@{}}
\toprule
\textbf{Outcome} & \textbf{Candidates} & \textbf{Sources} & \textbf{Pooled median [Q1, Q3]} & \textbf{Equal-source mean} \\
\midrule
Class-preserved & 699,250 & 100 & 0.6464 [0.6130, 0.6726] & 0.5719 \\
Adverse & 5,157 & 13 & 0.7039 [0.6822, 0.7191] & 0.7007 \\
Corrective & 9,031 & 9 & 0.6635 [0.6464, 0.6793] & 0.6751 \\
Lateral & 11,632 & 8 & 0.6567 [0.6397, 0.6774] & 0.6683 \\
\bottomrule
\end{tabular}
\end{table}

Class-changing candidates also showed larger source-matched internal divergence than preserved candidates. For adverse and corrective transitions, the relative $L_2$ contrast was positive at all seven boundaries after false-discovery-rate correction. For lateral transitions, six of seven boundaries were positive after correction; the stem was the exception ($q=0.1749$). Table \ref{tab:trace_contrasts} reports the complete set of mean contrasts and adjusted values.

\begin{table}[!htbp]
\centering
\caption{Source-matched transition-minus-preserved contrasts in relative $L_2$ activation change. Each cell gives the mean contrast above its adjusted $q$ value in parentheses. Source counts are 13 adverse, 9 corrective, and 8 lateral.}
\label{tab:trace_contrasts}
\small
\setlength{\tabcolsep}{3pt}
\renewcommand{\arraystretch}{1.15}
\begin{tabularx}{\columnwidth}{@{}>{\raggedright\arraybackslash}Xrrr@{}}
\toprule
\textbf{Boundary} & \textbf{Adverse} & \textbf{Corrective} & \textbf{Lateral} \\
\midrule
Stem        & \shortstack[r]{0.0168\\(0.0018)} & \shortstack[r]{0.0200\\(0.0230)} & \shortstack[r]{0.0098\\(0.1749)} \\
\addlinespace[3pt]
Attention 1 & \shortstack[r]{0.1815\\(0.0014)} & \shortstack[r]{0.1562\\(0.0072)} & \shortstack[r]{0.1535\\(0.0182)} \\
\addlinespace[3pt]
Local 1     & \shortstack[r]{0.1246\\(0.0014)} & \shortstack[r]{0.1179\\(0.0072)} & \shortstack[r]{0.1408\\(0.0092)} \\
\addlinespace[3pt]
Block 1     & \shortstack[r]{0.0844\\(0.0014)} & \shortstack[r]{0.0758\\(0.0072)} & \shortstack[r]{0.0827\\(0.0113)} \\
\addlinespace[3pt]
Attention 2 & \shortstack[r]{0.1528\\(0.0014)} & \shortstack[r]{0.1583\\(0.0081)} & \shortstack[r]{0.2164\\(0.0092)} \\
\addlinespace[3pt]
Local 2     & \shortstack[r]{0.1356\\(0.0014)} & \shortstack[r]{0.1436\\(0.0072)} & \shortstack[r]{0.1941\\(0.0102)} \\
\addlinespace[3pt]
Block 2     & \shortstack[r]{0.1046\\(0.0014)} & \shortstack[r]{0.1051\\(0.0072)} & \shortstack[r]{0.1404\\(0.0107)} \\
\bottomrule
\end{tabularx}
\end{table}

At the final block, the adverse contrast was 0.1046 (95\% source-bootstrap interval [0.0860, 0.1235]), the corrective contrast was 0.1051 ([0.0815, 0.1315]), and the lateral contrast was 0.1404 ([0.0651, 0.2668]).

The global and local paths changed differently across depth. For class-preserved candidates, attention-minus-local relative $L_2$ change was negative in block 1 ($-0.0477$, $q=0.00040$) but positive in block 2 (0.1281, $q=0.00040$). At block 2, the same contrast was positive for adverse (0.1194, $q=0.0069$), corrective (0.1085, $q=0.0195$), and lateral (0.0968, $q=0.0166$) candidates. None of the corresponding block-1 contrasts for class-changing outcomes survived correction. These comparisons are observational. They show where clean and perturbed computations differ, but they do not show which path controls the decision.

\subsection{Internal divergence and decision-score displacement}
\label{sec:norm_margin_results}

The source-level analysis shows that the magnitude of an internal response and its expression at the decision output are different quantities. Across all neighbors, the median within-source Spearman correlation of final-block relative $L_2$ change is 0.4410 with score-vector displacement and 0.3447 with absolute ground-truth margin change. When the 81,309 exact-zero responses are excluded, these medians fall to 0.2212 and 0.0544 across the same 100 sources (Table~\ref{tab:norm_margin}). The median correlation with signed margin change is 0.0078, with positive and negative source-specific relationships. Restricting further to preserved decisions with nonzero activation change gives a similar absolute-margin correlation of 0.0566.

\begin{table}[!htbp]
\centering
\caption{Median within-source Spearman correlations between final-block relative $L_2$ change and output changes. Each row summarizes 100 sources; values are descriptive medians.}
\label{tab:norm_margin}
\small
\begin{tabularx}{\columnwidth}{@{}>{\raggedright\arraybackslash}Xrrr@{}}
\toprule
\textbf{Candidates} & \shortstack{Score\\$L_2$} & \shortstack{Absolute\\margin} & \shortstack{Signed\\margin}\\
\midrule
All neighbors & 0.4410 & 0.3447 & 0.0032\\
Nonzero internal response & 0.2212 & 0.0544 & 0.0078\\
Preserved, nonzero response & 0.2206 & 0.0566 & 0.0138\\
\bottomrule
\end{tabularx}
\end{table}

Shared zero-change candidates contribute substantially to the association measured over the full neighborhood. Among actual nonzero responses, the activation norm only weakly ranks margin displacement. Recording the scores therefore adds decision-relevant information that cannot be replaced by the magnitude of the internal activation change alone.

\subsection{Clean internal states frequently restored adverse decisions}

The positive-control interventions behaved as required. Replacing the stem, block-1 output, or block-2 output with its matched clean tensor reproduced the clean score vector exactly and restored the clean prediction in 100\% of class-changing candidates.

For adverse candidates, replacement at an attention or local-path boundary often restored the correct clean decision. Table \ref{tab:adverse_patching} reports candidate-pooled and equal-source results. Their difference is expected because adverse candidates were concentrated in a small number of sources.

\begin{table}[!htbp]
\centering
\caption{Restoration of the clean, correct prediction after matched clean-activation replacement in adverse candidates. Equal-source intervals use 10,000 source-bootstrap samples.}
\label{tab:adverse_patching}
\small
\setlength{\tabcolsep}{3pt}
\renewcommand{\arraystretch}{1.15}
\begin{tabularx}{\columnwidth}{@{}>{\raggedright\arraybackslash}Xrr@{}}
\toprule
\textbf{Patched boundary} & \shortstack[r]{\textbf{Candidate-pooled}\\\textbf{rate}} & \shortstack[r]{\textbf{Equal-source mean}\\\textbf{{[95\% interval]}}} \\
\midrule
Attention 1 & 77.18\% & \shortstack[r]{92.93\%\\{[86.45, 98.24]}} \\
\addlinespace[3pt]
Local 1     & 71.22\% & \shortstack[r]{92.31\%\\{[83.60, 99.29]}} \\
\addlinespace[3pt]
Attention 2 & 77.64\% & \shortstack[r]{91.26\%\\{[83.56, 97.82]}} \\
\addlinespace[3pt]
Local 2     & 41.17\% & \shortstack[r]{78.51\%\\{[61.79, 92.76]}} \\
\bottomrule
\end{tabularx}
\end{table}

The largest raw attention-minus-local restoration contrast occurred in block 2 for adverse candidates. The source-equal difference was 0.1276, with a bootstrap interval of [0.0210, 0.2580] and unadjusted $p=0.0486$. It did not survive correction within the prespecified patch family ($q=0.2916$). No attention-versus-local patch contrast survived false-discovery-rate correction. The results support clean-boundary sufficiency but not a claim that either branch is causally dominant.

\begin{figure}[!htbp]
\centering
\includegraphics[width=\textwidth]{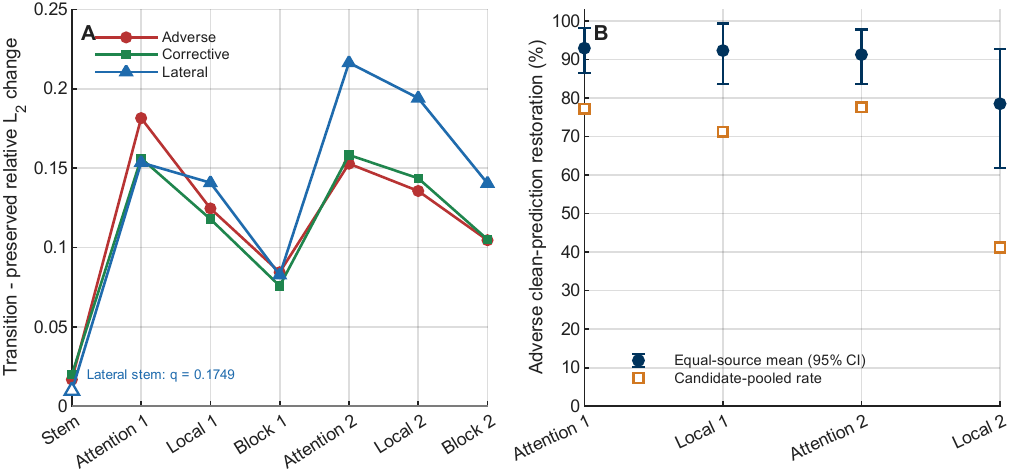}
\caption{Internal divergence and matched clean-activation replacement. Panel A shows source-matched transition-minus-preserved contrasts in relative $L_2$ activation change across seven internal boundaries. The lateral contrast at the stem is the only plotted transition contrast that did not survive false-discovery-rate correction. Panel B shows restoration of the clean, correct prediction for adverse candidates after replacing the candidate activation with the matched clean activation at the four branch boundaries. Error bars show the 95\% source-bootstrap intervals of the equal-source means. The candidate-pooled rates are shown separately because adverse candidates are strongly concentrated by source.}
\label{fig:internal_recovery}
\end{figure}

Figure \ref{fig:internal_recovery} combines the observational and intervention results. Panel A shows the excess relative $L_2$ change of each transition type over source-matched preserved candidates across the seven boundaries. Panel B shows adverse clean-prediction restoration at the four branch boundaries using candidate-pooled and equal-source summaries.

For corrective candidates, restoring the clean state tends to recover the original incorrect prediction. Across the four branch boundaries, the equal-source clean-prediction restoration rate ranged from 67.70\% to 78.63\%, whereas ground-truth restoration ranged from 15.09\% to 24.64\%. For lateral candidates, clean-prediction restoration ranged from 72.15\% to 78.82\%, while ground-truth restoration remained between 0.22\% and 0.92\%. These are intervention outcomes, not repair rates.

\subsection{Exact temporal aggregation in two count-readout replicas}
\label{sec:count_readout_results}

The complementary audit completed 1,297,705 neighbors for each replica, giving 2,595,410 model-neighbor evaluations on the same 100 utterance identities. Table~\ref{tab:count_readout} gives the complete trace-readout-decision partition. Most candidates changed the count representation while preserving the class (T2). A further 97,293 and 77,753 candidates changed internal spikes while preserving the count vector (T1).

\begin{table}[!htbp]
\centering
\caption{Complete outcomes in two separately seeded count-readout SNN replicas. Each column contains 1,297,705 unique neighbors of the same 100 sources. The final-layer row is a subset of T1, adverse is a subset of T3, and a robust source is clean-correct for every allowed neighbor.}
\label{tab:count_readout}
\small
\begin{tabularx}{\columnwidth}{@{}>{\raggedright\arraybackslash}Xrr@{}}
\toprule
\textbf{Outcome} & \textbf{Replica A} & \textbf{Replica B}\\
\midrule
T0: trace unchanged & 174,198 & 168,422\\
T1: trace changed, counts preserved & 97,293 & 77,753\\
T2: counts changed, class preserved & 1,022,650 & 1,032,964\\
T3: class changed & 3,564 & 18,566\\
\addlinespace[3pt]
Final-layer trace changed, counts preserved & 89,176 & 72,380\\
\midrule
Adverse neighbors (subset of T3) & 790 & 15,773\\
Initially correct sources (of 100) & 73 & 75\\
Correct and robust sources (of 100) & 67 & 70\\
\bottomrule
\end{tabularx}
\end{table}

Of the T1 outcomes, 89,176 and 72,380 involve a changed final-layer spike tensor with exactly preserved global and window counts. Such reachable input-induced changes constitute 7.94\% and 6.41\% of all changed traces, respectively. Both replicas exhibit cases in all 100 utterances. The equality of the readout for any $P\Delta S=0$ is a property of the architecture; the frequency with which allowed inputs reach that nullspace is the empirical observation. The remaining 8,117 and 5,373 T1 outcomes (0.72\% and 0.48\% of changed traces) involve earlier changes that disappear before the final spike layer. The audit thus separates changes removed before the readout, changes removed by temporal aggregation, and changes retained in the count representation without crossing the decision boundary.

The paired replicas expose a second form of aggregation ambiguity. Their validation accuracies differ by only 0.2505 percentage points (72.2773\% versus 72.5278\%), yet B has 5.21 times as many class-changing neighbors as A on the same 100 identities and 1,297,705 neighbors per model. This gap is concentrated in four sources (914, 4499, 7289, and 8134) that are clean-correct only in B: their T3 counts are 369 in A and 15,760 in B. Across the other 96 sources, the counts are 3,195 versus 2,806, so the full-panel gap does not represent uniformly greater vulnerability in B. A paired, within-label source bootstrap gives a B-minus-A equal-source class-change rate difference of $+1.1896$ percentage points (95\% interval [0.0230, 2.4507]). There are 67 sources in A and 70 in B that are clean-correct and remain correct for every tested neighbor. Among the 67 sources clean-correct in both models, A has 248 adverse neighbors and B has 13. The four sources also account for 15,760 of B's 15,773 adverse neighbors. The full-panel B-minus-A adverse-rate difference is $+1.3001$ percentage points [0.1843, 2.5593], whereas on the common-correct cohort it is $-0.0417$ [${-}0.0923$, 0.0003]; these are distinct conditioned comparisons. Adverse neighbors occur for 6 of 73 and 5 of 75 initially correct sources, respectively. The replicas use their native 10 ms neighborhoods and are not pooled with SpikeSCR. Their paired result shows that accuracy, candidate-level class-change load, and fully robust source counts can rank the same two fixed models differently.

\subsection{Temporal transitions persist across CPU and GPU execution}
\label{sec:device_persistence}

The paired device replay completed all 25,820 originally class-changing candidates. GPU predicted classes reproduce the canonical map without a single disagreement. CPU and GPU agree on 25,794 candidates (99.8993\%) and on transition type for 25,795 (99.9032\%). Of 5,157 canonical adverse moves, 5,152 remain adverse on CPU (99.9030\%); one of these changes its incorrect target class. Five adverse, nine corrective, and eleven lateral candidates return to their clean predicted class on CPU (Table~\ref{tab:device_replay}). The median source-candidate maximum absolute score difference is $1.43\times10^{-6}$, the 95th percentile is $2.86\times10^{-6}$, and the maximum is 1.1871. All 100 clean panel labels agree across devices.

Both devices recover the same 13 adverse-sensitive sources, and all 13 preselected first adverse witnesses retain the same incorrect class on both devices. The concentrated temporal vulnerability is therefore reproducible across these CPU and GPU executions, rather than confined to one device. This replay tests persistence of the original class-changing cohort; it does not recalculate vulnerability incidence over the originally preserved candidates.

\section{Discussion}
\label{sec:discussion}

\subsection{What average performance leaves unresolved}

The central result is a mismatch between favorable aggregate behavior and local source vulnerability. On the fixed SpikeSCR panel, equal-source expected accuracy under a uniform draw from each stated neighborhood improves, yet 13 initially correct utterances have an adverse neighbor. Corrective transitions offset adverse transitions in the average. The complete map resolves both effects and measures what informed as well as uniform searches miss. The paired replicas provide a related warning: near-identical validation accuracies coexist with a 5.21-fold difference in class-changing candidates driven by four sources, while the ordering of fully robust correct sources runs the other way.

The experiments resolve two different losses of information when computations are compressed to labels and then to average accuracy. A favorable mean can coexist with adverse moves concentrated in particular inputs. A preserved class can accompany a median final-block relative $L_2$ change of 0.6464 in SpikeSCR; these residual block outputs are continuous-valued and this number is not a spike-flip rate. The count-readout replicas additionally exhibit changed final-layer spike timing with exactly preserved readout counts. Decision stability, representation stability, and trajectory stability therefore have different measurements.

A larger activation norm difference is not itself evidence of degraded computation. Both adverse and corrective transitions show increased source-matched divergence. Among nonzero final-block responses, the median within-source correlation with absolute ground-truth margin movement is only 0.0544. Its direction relative to the ground truth must therefore be retained. The count-readout audit supplies a concrete aggregation mechanism for its own architecture, while the SpikeSCR interventions test the sufficiency of clean branch states in the global-local network.

\subsection{Why complete local coverage changes the assessment}

Optimized attacks seek informative harmful examples, and formal verification addresses bounded existence questions. The present audit retains the full outcome distribution after the first witness. Low-margin allocation detects more vulnerable sources in expectation than uniform sampling on this panel, and a single-gradient ranking finds a different set of rare sources. No search wins on every source: hard low-margin focus excludes source 1627 at the reported budgets, while the gradient top-1,000 misses sources 9282 and 1534. Their adverse rates are one or two moves among thousands. A post hoc longer gradient prefix finds all 13, so the case for complete enumeration rests on full distributional measurement and verified coverage within the defined neighborhood, not on a claim that witness discovery always requires exhaustive work.

The analysis also separates the unit of perturbation from the unit of inference. Candidate-pooled rates characterize distinct input changes; equal-source summaries characterize the fixed label-quota panel. Source-level resampling respects the dependence among moves from the same utterance. The exact coverage calculation then uses each source's full neighborhood size and adverse count, with no assumption that all utterances share one adverse probability.

Clean margin supplies useful predictive information, but it does not determine every outcome. Normalizing by the source horizon leaves the lowest-quartile allocation unchanged in this panel. Its negative association with adverse rate agrees with a decision-boundary interpretation, while an excluded vulnerable source shows why margin is not a certificate. Margin-based robustness guarantees additionally require control of score variation with the input \citep{hein2017}; this audit measures actual finite-neighborhood outcomes. The tested global temporal contrasts did not establish a consistent pattern after correction.

\subsection{Matched interventions establish clean-state sufficiency}
The trace results and the interventions answer different questions. Trace divergence is observational: it measures where clean and retimed runs differ. Matched replacement is interventional: it asks whether supplying the clean state at a declared boundary is sufficient to recover a downstream decision. The adverse-recovery rates show that a clean branch state often carries enough information to restore the correct decision. Because adverse candidates cluster by source, candidate-pooled rates (41.17 - 77.64\%) are lower than equal-source means (78.51 - 92.93\%) across the four branch boundaries; both views are needed to characterize the intervention.

The combination of trace and replacement results is particularly informative. At block 2, attention showed greater relative divergence than the local path, yet no attention-versus-local recovery contrast survived multiple-testing correction. The observed difference in activation sensitivity was not accompanied by an established difference in recovery effects. Nonsignificant recovery contrasts do not establish equivalence between the branches. The available evidence supports recovery by matched clean states at several boundaries, without establishing a universally dominant branch. The full-stem and full-block replacements restored 100\% by design and verified the intervention pipeline as positive controls.

Corrective transitions further show why the target of restoration must be stated. In an initially incorrect source, restoring the clean prediction means restoring an error. A patch may therefore recover the original computation while reducing task correctness. For this reason, clean-prediction and ground-truth restoration were kept separate, and the main recovery interpretation was restricted to adverse candidates.

\subsection{Implications for temporal robustness}
The audit suggests a practical measurement protocol for future models. A temporal-robustness report should define the operator exactly; distinguish unique inputs from event multiplicity; retain transition direction; report both candidate-pooled and source-equal rates; identify the fraction of sources with any adverse neighbor; and show how adverse load is concentrated. A single accuracy value or attack success rate does not contain this information.

The results also motivate, but do not test, several training strategies. Candidate neighborhoods could be sampled during training, low-margin sources could receive targeted temporal augmentation, and a consistency term could penalize excessive boundary-state change for label-preserving moves. Adverse-neighbor training could also be compared with random jitter. These are hypotheses for future experiments. This work evaluates a frozen classifier and does not claim that any of these procedures will improve robustness without a separate training study and untouched evaluation data.

\subsection{Execution dependence has an identifiable mechanism}

The controlled execution results expose another limitation of aggregate reporting. Reversing batch order changes 531 validation decisions although the net accuracy shift is only 0.4709 percentage points; score vectors change for every source. Source isolation restores exact order invariance and reproduces padding-matched singleton labels on 9,980 of 9,981 inputs. The public q/k reshape sends different sources through shared temporal LIF lanes, and isolating those lanes removes the measured order effect. The checkpoint was trained with batch size 256 and the same public reshape, so its weights were optimized under the coupled training implementation rather than a singleton or source-isolated path. The observed accuracy ordering (ascending-ID batches 86.3340\%, isolated batches 86.0535\%, reversed batches 85.8631\%) shows sensitivity to arbitrary batch context. It does not establish that neighboring sources provide task-relevant information. The execution contract must therefore specify both training and evaluation paths. These controls concern our checkpoint and the pinned implementation; the execution path behind the published SpikeSCR results was not independently established.

Falcon's hardware latency model reconstructs SpikeSCR using per-head query and key matrices with rotary encoding followed by a second LIF stage \citep{yan2026falcon}. That logical description does not establish a running implementation's batched storage semantics. Our controls show why hardware scheduling assumptions should be checked against the actual time and source axes.

Padding has a different diagnosis. Its readout contains extra softmax mass, but excluding that mass restores only 26 of 344 changed labels; the other 318 already differ within the valid output prefix. The \texttt{v2} attention mask zeros invalid key columns of $QK^{\mathsf T}$ before multiplication by $V$. It does not mask the preceding q/k LIF computations, the local convolution, or the final temporal readout. Horizon-dependent q/k reshaping and propagation of padded-position activity through the symmetric 31-tap local convolution are candidate routes for the prefix effect; their individual contributions were not isolated. Device persistence answers a separate question: GPU replay exactly reproduces the canonical class-changing predictions, while CPU replay preserves all 13 vulnerable sources and their fixed witnesses.

\subsection{Limitations}

The detailed global-local trace and intervention analysis concerns one SpikeSCR initialization and checkpoint. The count-readout comparison adds two seeds in a second, less accurate architecture with a different native temporal resolution and readout. It establishes neither quantitative replication of the SpikeSCR rates nor that spiking dynamics cause the observed mismatch relative to a nonspiking model. All local analyses use the same 100 label-stratified validation identities, including only 13 SpikeSCR adverse-sensitive sources. Source count, rather than the hundreds of thousands of dependent candidates, limits population precision; the source-bootstrap adverse-rate interval is wide (0.25 - 1.70\%). These source-equal summaries describe the fixed panel, not prevalence over all 9,981 validation utterances or training seeds.

The perturbations are adjacent-bin, same-feature one-count moves at a fixed horizon in the transformed 140-feature input. Their physical or perceptual equivalence to sensor jitter is not established; a real timing change may cross no bin boundary or move multiple integrated counts. The audit does not cover larger budgets or unrestricted temporal robustness. Margin policies were specified retrospectively on the observed panel, and the gradient comparisons use fixed first-order estimates, including all 34 rival classes, rather than iterative optimized attacks. Their candidate-query counts omit gradient computation. The new full-validation singleton run and clean gradients required additional model computation, whereas the sampling expectations and paired replica summaries reuse completed maps.

The batch-256 benchmark and singleton local audit have distinct execution contracts. Controlled q/k source isolation identifies the mechanism of the measured batch-order effect; it does not isolate every cause of the remaining padding or batch-shape differences. These findings concern the pinned public implementation and recorded checkpoint. A layout-corrected q/k path that explicitly permutes head and time axes would define a third function for which the weights were not optimized; that function was not evaluated. The archived test result belongs to batch-256 execution; no singleton or source-isolated test measurement was made. The device replay covers all originally class-changing neighbors, not the 699,250 originally preserved ones, and therefore does not estimate new adverse incidence outside that cohort. Matched clean-activation replacement assumes access to the clean state and establishes sufficiency under the stated intervention, not a deployed defense or proof of unique causal necessity.

\section{Conclusion}
Complete temporal auditing exposes the source distribution hidden by aggregate performance. On the SpikeSCR panel, equal-source expected accuracy under a uniformly selected one-count neighbor increases from 84.00\% to 84.54\%, while 13 of 84 initially correct utterances admit adverse neighbors and five carry 93.21\% of harmful candidates. Low-margin sampling improves expected discovery, and a fixed gradient ranking reaches different rare cases. The complete retained map establishes every transition and each source's adverse density, including those unseen at the compared query budgets.

Internal measurements extend this account beyond the output label. Source-matched traces distinguish activation magnitude from harmful direction, and clean branch states often restore correct adverse predictions. In a second architecture, similarly accurate count-readout replicas have a 5.21-fold class-change gap concentrated in four sources, with the opposite ordering in fully robust correct source counts. Controlled q/k isolation identifies and removes the 531-label batch-order effect while preserving singleton processing. Paired CPU/GPU replay reproduces the same 13 vulnerable sources and their preselected witnesses, with 99.8993\% candidate-label agreement. These findings support reporting source incidence, candidate density, internal response, and the execution contract alongside aggregate accuracy.

\section*{Data and code availability}
The SSC dataset is available through the Heidelberg Spiking Data Sets at
\url{https://zenkelab.org/resources/spiking-heidelberg-datasets-shd/}.
The original SpikeSCR code is available at
\url{https://github.com/JackieWang9811/SpikeSCR}.
Code, the frozen checkpoint, the complete outcome map of the 725,070 neighbors, the internal activation metrics, the 180,740 replacement score vectors, the search-policy records, the replica records and the execution-control records are archived at Zenodo (\url{https://doi.org/10.5281/zenodo.23089436}) and maintained at \url{https://github.com/DrCanD/spikescr-local-event-geometry}. The command \texttt{python -m ssc\_geometry paper} regenerates every table of this article from the archived records without model inference and compares 434 printed values with their regenerated counterparts. Raw hidden tensors, unpatched census score vectors, full all-target gradient tensors and the replica weights are not included.

\FloatBarrier
\clearpage
\begingroup
\urlstyle{same}

\endgroup
\FloatBarrier

\clearpage
\appendix
\numberwithin{equation}{section}
\numberwithin{table}{section}
\numberwithin{figure}{section}
\section*{APPENDIX}
\section{Numerical execution and canonical reconstruction}
\label{app:execution}

The archived batch-locality controls identified deterministic score dependence on sample position in the pinned rotary-reshape-LIF q/k path. A samplewise adapter was introduced to reproduce the public singleton mapping independently for each sample. A batch-parity check during development recorded prediction parity for 32 candidates and a maximum adapter–singleton score difference of $5.7220458984375\times10^{-6}$; later internal checks nevertheless found rare score disagreements. The identical rounded maximum in the independent CPU-panel comparison was recomputed from its own score arrays and occurs at source 6260, class 19. These are distinct comparisons, not a reused measurement. The subsequent 51,560-candidate canonicality audit recorded nine transition disagreements and required a complete singleton reconstruction. The precise numerical source of the residual adapter-singleton differences was not isolated.

The complete reconstruction evaluated all 725,070 candidates with the unmodified network at computational batch size one. It corrected ten candidate predictions across six sources relative to the rejected adapted executor. The final class-changing count was 25,820, partitioned into 5,157 adverse, 9,031 corrective, and 11,632 lateral outcomes. The internal bridge then retained only singleton traces with scores equal to these canonical outcomes and completed the remaining singleton traces and interventions. The earlier adapted outcome map is not used in any result reported here.

This correction establishes the execution provenance of the finite audit. The ten-candidate comparison concerns the rejected adapted executor and canonical CUDA singleton inference. The later full-validation CPU run evaluates a CPU singleton path that differs from the canonical GPU path on two labels (Table~\ref{tab:execution_controls}). Its 8,591 correct predictions and 617 disagreements with archived batch-256 execution describe that separate CPU control. Within the 100-source panel, canonical GPU and CPU singleton labels agree; canonical GPU and archived batch-256 labels disagree on eight sources, including two that are adverse-sensitive under the singleton audit but initially wrong under batch-256 execution. The archived test result belongs to its recorded batch-256 path.

\subsection{Controlled full-validation GPU replays}
The frozen validation cache, epoch-282 checkpoint, pinned source files, and CPU singleton reference were hash-verified before the GPU controls. PyTorch 2.11.0 with CUDA 12.8 and cuDNN 9.19 ran on an NVIDIA RTX PRO 6000 Blackwell Server Edition. Native singleton inputs retain their own horizons. The padding-matched singleton condition uses zero padding to the maximum native horizon of the source's contiguous batch-256 group, with the same valid-input mask. In the checkpoint's \texttt{v2} attention branch, this mask is expanded over the key axis of $QK^{\mathsf T}$ and sets invalid key columns to zero before multiplication by $V$. It is passed to the attention module in each block, but does not gate the embedding, q/k LIF stages, local convolution, or readout. The public readout sums softmax values over all padded output bins in both conditions; no new output masking is introduced. Sources are grouped by ascending validation identifier, with the final group retaining its size of 253. The reversed condition reverses source order within each otherwise unchanged padded group and restores the original identifier order before comparison.

All four conditions use the same model and GPU runtime, float32, evaluation mode, disabled automatic mixed precision and TF32, deterministic algorithms, and a state reset at each forward pass. Each of ten selected native singleton inputs was replayed with exact score equality required. The native GPU singleton scores match the 100 canonical panel score vectors bit-for-bit. All input identities, labels, horizons, and hashes match the frozen CPU reference. The complete recorded arrays independently reproduce every accuracy and paired count in Table~\ref{tab:execution_controls}.

\begin{table}[!htbp]
\centering
\caption{Paired execution controls on all 9,981 frozen validation sources. Each row compares condition A with condition B. C$\to$W counts correct-to-wrong predictions; W$\to$C counts wrong-to-correct predictions; W$\ne$W counts cases wrong in both conditions with different predicted classes. Changed labels equal the sum of these three counts. The q/k-isolated row uses the source-isolation intervention defined below. The last row checks the archived batch-256 labels against their fresh replay.}
\label{tab:execution_controls}
\small
\begin{tabularx}{\textwidth}{@{}Xrrrrr@{}}
\toprule
\textbf{Condition A $\to$ condition B} & \textbf{Accuracy A $\to$ B (\%)} & \textbf{Changed labels} & \textbf{C$\to$W} & \textbf{W$\to$C} & \textbf{W$\ne$W} \\
\midrule
CPU singleton $\to$ native GPU singleton & 86.0735 $\to$ 86.0836 & 2 (0.02\%) & 0 & 1 & 1 \\
Native $\to$ padding-matched GPU singleton & 86.0836 $\to$ 86.0435 & 344 (3.45\%) & 110 & 106 & 128 \\
Padding-matched GPU singleton $\to$ GPU batch 256 & 86.0435 $\to$ 86.3340 & 614 (6.15\%) & 191 & 220 & 203 \\
Native GPU singleton $\to$ GPU batch 256 & 86.0836 $\to$ 86.3340 & 619 (6.20\%) & 199 & 224 & 196 \\
GPU batch 256 $\to$ reversed GPU batch 256 & 86.3340 $\to$ 85.8631 & 531 (5.32\%) & 205 & 158 & 168 \\
Padding-matched singleton $\to$ q/k-isolated batch 256 & 86.0435 $\to$ 86.0535 & 1 (0.01\%) & 0 & 1 & 0 \\
q/k-isolated batch 256 $\to$ reversed & 86.0535 $\to$ 86.0535 & 0 (0.00\%) & 0 & 0 & 0 \\
Archived $\to$ fresh GPU batch 256 & 86.3340 $\to$ 86.3340 & 0 (0.00\%) & 0 & 0 & 0 \\
\bottomrule
\end{tabularx}
\end{table}

Only sources 7368 and 7558 change class between CPU and native GPU singleton execution. For each source, let $\delta_i=\max_c|s^{\mathrm{CPU}}_{ic}-s^{\mathrm{GPU}}_{ic}|$. All 9,981 values are nonzero; their median is $1.90735\times10^{-6}$, 95th percentile $3.81470\times10^{-6}$, and maximum 3.0513. There are 134 sources with $\delta_i>10^{-4}$. The bulk of the score differences is therefore small despite the larger tail. The padding and batched-execution contrasts remain distinct from this device contrast and cannot be summed to partition the 619 native-singleton versus batch-256 label disagreements. The additional q/k study exactly reproduces all original and reversed batch-256 labels before intervening.

\subsection{Padding gap and output decomposition}
\label{app:padding_decomposition}

Let $H_g$ be the maximum native horizon in a source's batch group and let $p_{itc}^{\mathrm{pad}}$ be its padded-run softmax output. The public readout decomposes as
\begin{equation}
\begin{aligned}
s_{ic}^{\mathrm{pad}}&=s_{ic}^{\mathrm{prefix}}+s_{ic}^{\mathrm{extra}},\\
s_{ic}^{\mathrm{prefix}}&=\sum_{t=0}^{H_i-1}p_{itc}^{\mathrm{pad}},\qquad
s_{ic}^{\mathrm{extra}}=\sum_{t=H_i}^{H_g-1}p_{itc}^{\mathrm{pad}}.
\end{aligned}
\label{eq:padding_decomposition}
\end{equation}
Consequently $\sum_c s_{ic}^{\mathrm{extra}}=H_g-H_i$. The valid-prefix output is taken from the padded run; it need not equal the native-horizon output. Native versus prefix scores, prefix versus full predictions, and the extra-output vector were retained for every diagnostic source. These identities were checked numerically before aggregation. Table~\ref{tab:padding_gap} reports the full-validation association between added bins and changed labels. It is descriptive, since padding gap was not randomly assigned.

\begin{table}[!htbp]
\centering
\caption{Native versus padding-matched GPU singleton labels over all 9,981 validation sources, grouped by the number of added 5 ms bins.}
\label{tab:padding_gap}
\small
\begin{tabular}{@{}lrrr@{}}
\toprule
\textbf{Added bins} & \textbf{Sources} & \textbf{Changed} & \textbf{Rate (\%)}\\
\midrule
0 & 4,057 & 0 & 0.00\\
1 - 5 & 5,004 & 236 & 4.72\\
6 - 10 & 45 & 2 & 4.44\\
11 - 20 & 155 & 13 & 8.39\\
21 - 40 & 294 & 26 & 8.84\\
41 - 80 & 322 & 43 & 13.35\\
81 or more & 104 & 24 & 23.08\\
\bottomrule
\end{tabular}
\end{table}

All 344 historically changed sources and 64 positive-padding unchanged controls were rerun. Their native and full padded labels exactly reproduced the stored references. Removing the added-bin component restores 26 changed labels, whereas 318 remain changed in the prefix alone. Holding native output fixed and adding the observed extra-output vector changes 43 labels in this selected cohort; this algebraic diagnostic is not a separate physical perturbation. Both prefix alteration and added output mass must therefore be retained in the explanation. Zero input at a padded position does not guarantee zero downstream activity after affine normalization and stateful spiking operations. The actual attention mask already excludes invalid key columns from the attention product, so applying the same post-LIF k/v mask would duplicate that exclusion. The horizon-dependent public q/k reshapes can nevertheless alter LIF sequences before masking, and the symmetric local kernel can carry padded-position activity into valid positions. The 4.72\% label-change rate with only one to five added bins is consistent with a prefix effect, but does not identify either route as its unique cause. No branch-specific padding intervention was performed.

\subsection{Source isolation in the q/k LIF layout}
\label{app:qk_mechanism}

In the pinned attention implementation, rotary-encoded queries and keys are contiguous in $[B,h,T,d]$ layout and are directly reshaped to $[T,B,N]$ before \texttt{q\_lif2} or \texttt{k\_lif2}, with $N=hd$. This reshape reinterprets the axis meanings without preserving source membership along each declared LIF time sequence. In the recorded $T=200$, $B=256$, $N=256$ group, each of the 65,536 LIF membrane lanes receives entries owned by 200 different original sources across its declared time axis. Reversing the source order changes those sequences.

For either q/k branch, let $R$ denote the caller's $[T,B,N]$ tensor after the public reshape. The diagnostic wrapper forms independent sequences and restores the caller's expected layout:
\begin{equation}
\begin{aligned}
U&=\operatorname{permute}_{(1,0,2)}
   \!\left(\operatorname{reshape}_{(B,T,N)}R\right),\\
Y&=\operatorname{LIF}(U),\\
R'&=\operatorname{reshape}_{(T,B,N)}
   \!\left(\operatorname{contiguous}
   \!\left(\operatorname{permute}_{(1,0,2)}Y\right)\right).
\end{aligned}
\label{eq:qk_source_isolation}
\end{equation}
Contiguous storage is also enforced after forming $U$. This retains the exact public singleton head/time flattening convention for each source. It does not introduce a conventional temporal-axis correction or retrain the model. For $B=1$, the original LIF forward is called directly, so the unchanged singleton scores are a bypass invariant. The batch transformation is assessed by the full 9,981-source comparison and the pair controls below. Wrappers are restored after every diagnostic condition and are inactive during canonical candidate replay.

The three pairs were selected deterministically from the first three historical order-sensitive batch groups. Original and q-only or k-only conditions show nonzero order and duplicate-input position score differences. Joint isolation removes both differences in every pair; the maximum discrepancy from padded singleton scores across these three pairs is $1.91\times10^{-6}$. The full-validation contrast keeps checkpoint, GPU, group composition, padded horizon, input mask, and batch/time shapes fixed within each paired order comparison. All 9,981 joint-isolation score vectors are bitwise order-invariant. Against the historical padding-matched singleton reference, isolated batch labels agree on 9,980 of 9,981 sources. Source 7942 changes from class 27 to its correct class 25. Across this full cross-shape comparison, source-wise maximum absolute score differences have median $1.91\times10^{-6}$, 95th percentile $3.81\times10^{-6}$, and maximum 3.7136; 54 sources exceed $10^{-4}$. The pair-control maximum must therefore not be interpreted as the full-validation maximum. Exact within-shape order invariance and near-complete cross-shape label agreement establish different properties.

\subsection{Paired native-singleton device replay}
\label{app:device_replay}

The 25,820 canonical class-changing tensors were reconstructed from retained source, candidate, from-bin, to-bin, and feature coordinates. Total count, non-negativity, native horizon, and candidate identities were checked. Each candidate was run once on the AMD EPYC 9B45 CPU and once on the NVIDIA RTX PRO 6000 Blackwell Server Edition in the same PyTorch 2.11.0/CUDA 12.8 runtime; CPU inference used one thread. Table~\ref{tab:device_replay} distinguishes retention of transition type from retention of the exact predicted class.

\begin{table}[!htbp]
\centering
\caption{CPU replay of every originally class-changing candidate. GPU retains all canonical types and predicted classes. ``Same type'' means adverse, corrective, or lateral remains the same under the device-specific clean prediction; ``same class'' means exact agreement with the canonical candidate label.}
\label{tab:device_replay}
\small
\begin{tabular}{@{}lrrr@{}}
\toprule
\textbf{Canonical type} & \textbf{Candidates} & \shortstack{\textbf{CPU: same}\\\textbf{type}} & \shortstack{\textbf{CPU: same}\\\textbf{class}}\\
\midrule
Adverse & 5,157 & 5,152 & 5,151\\
Corrective & 9,031 & 9,022 & 9,022\\
Lateral & 11,632 & 11,621 & 11,621\\
\midrule
Total & 25,820 & 25,795 & 25,794\\
\bottomrule
\end{tabular}
\end{table}

The 25 transition-type disagreements are returns to the clean class; the remaining label disagreement changes one adverse candidate's incorrect target while preserving its adverse type. The first archived adverse witness of each vulnerable source is unchanged across devices. The original 13 source identities are therefore recovered independently of selecting new device-specific witnesses. Full CPU/GPU score vectors, coordinates, witness identities, per-source counts, contract, and file hashes are retained. No originally class-preserved candidate is included in this selected-cohort replay.

\end{document}